\pdfoutput=1

\documentclass[11pt]{article}

\usepackage[final]{acl}

\usepackage{times}
\usepackage{latexsym}

\usepackage[T1]{fontenc}

\usepackage[utf8]{inputenc}

\usepackage{microtype}

\usepackage{inconsolata}

\usepackage{graphicx}
\usepackage{amsmath}
\usepackage{amsfonts}
\usepackage{comment}
\usepackage[export]{adjustbox}
\usepackage{tabularray}
\usepackage{array}
\newcolumntype{P}[1]{>{\centering\arraybackslash}p{#1}}

\usepackage{multirow}
\usepackage{pdflscape}
\usepackage[symbol]{footmisc}

\title{zFLoRA: Zero-Latency Fused Low-Rank Adapters}

\author{Dhananjaya Gowda$^*$ \qquad Seoha Song$^*$ \qquad Harshith Goka \qquad Junhyun Lee \\
Samsung Research \\
\texttt{\{d.gowda, seoha.song, h9399.goka, junhyun8.lee\}@samsung.com}\\}

\begin{document}
\maketitle

\begin{abstract}
Large language models (LLMs) are increasingly deployed with task-specific adapters catering to multiple downstream applications.
In such a scenario, the additional compute associated with these apparently insignificant number of adapter parameters (typically less than 1\% of the base model) turns out to be disproportionately significant during inference time (upto 2.5x times that of the base model).
In this paper, we propose a new {\bf zero-latency fused low-rank adapter (zFLoRA)} that introduces zero or negligible latency overhead on top of the base model.
Experimental results on LLMs of size 1B, 3B and 7B show that zFLoRA compares favorably against the popular supervised fine-tuning benchmarks including low-rank adapters (LoRA) as well as full fine-tuning (FFT).
Experiments are conducted on 18 different tasks across three different categories namely commonsense reasoning, math reasoning and summary-dialogue.
Latency measurements made on NPU (Samsung Galaxy S25+) as well as GPU (NVIDIA H100) platforms show that the proposed zFLoRA adapters introduce zero to negligible latency overhead. 
\end{abstract}

\footnotetext{$^*$ Equal contributions. Accepted at EMNLP-2025 (Main).}

\section{Introduction}
\label{sec:intro}
Large language models (LLMs) are increasingly becoming popular and are on their way to become an indispensable part of our day to day life~\cite{gemmateam2025gemma3technicalreport, dubey2024llama, openai2024gpt4_etal, deepseekai2025deepseekv3technicalreport_etal}.
The most powerful of these LLMs have several hundreds of billions of parameters and are often deployed on cloud computing services due to their high computational load.
However, the fast evolving techniques on model compression, quantization and other optimizations have made small to medium sized LLMs to catch up with their huger counterparts on a large subset of tasks that the LLMs can handle.
It has been shown that a small to medium sized LLM when fine-tuned using a small number of adapter parameters and task specific data can perform as good as a huge LLM~\cite{deepseekai2025deepseekv3technicalreport_etal, liu2024mobilellmoptimizingsubbillionparameter, allal2025smollm2smolgoesbig, dubey2024llama}.
In light of these developments, coupled with the concerns on data privacy and security, small to medium sized LLMs are increasingly being deployed on end-user devices such as mobiles, computers, robots, automobiles, etc., as well as other edge platforms and devices~\cite{xu2024ondevicelanguagemodelscomprehensive}.
\begin{figure}[t]
    \centering    
        \includegraphics[width=\columnwidth]{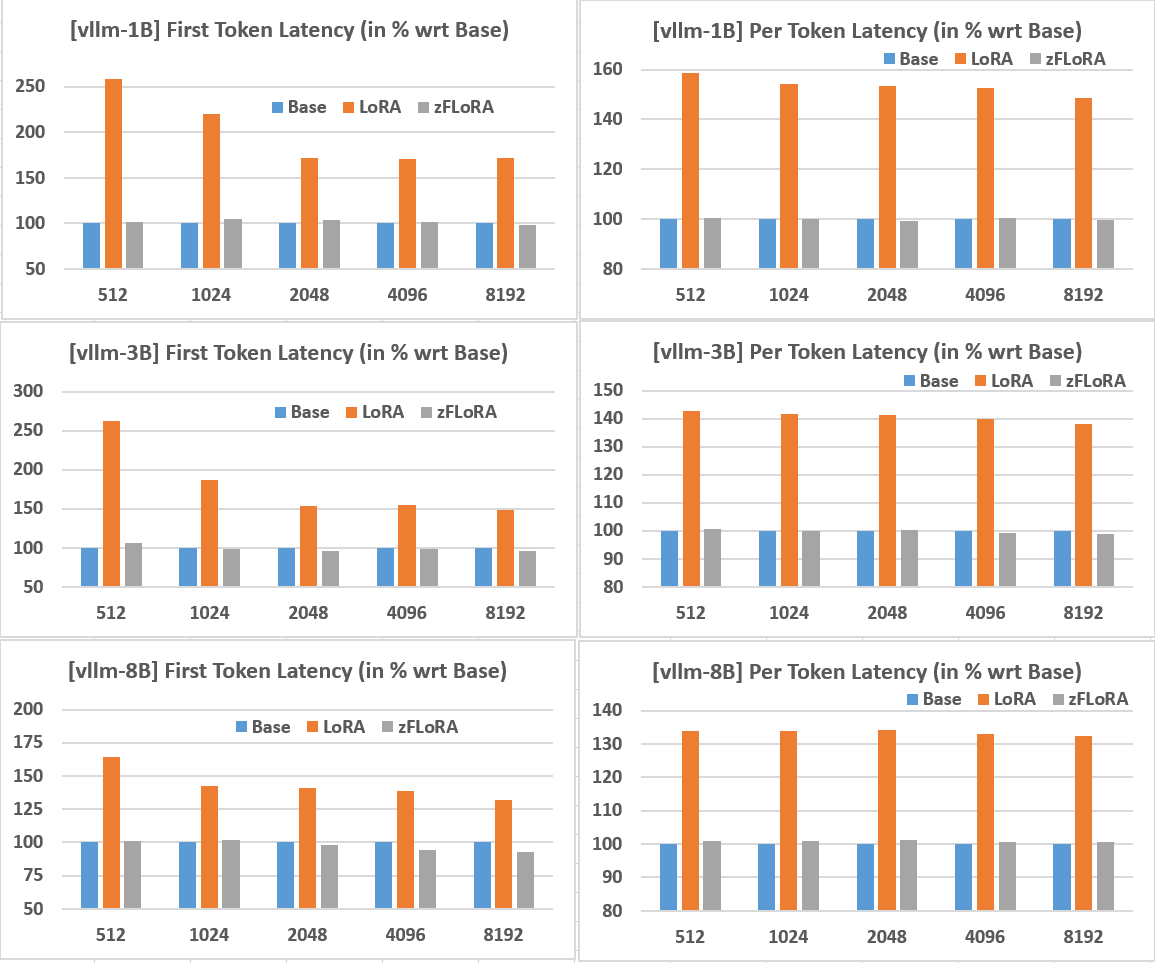}
    \caption{Inference latencies (first-token and per-token) of LoRA and zFLoRA for different input prompt lengths (512 to 2048) using vllm inference engine on NVIDIA H100 GPU at FP16 precision, expressed as a percentage of the base model (LLaMA 1B, 3B and 8B) latencies.}
    \label{fig:vllmlat}
    \vspace{-5mm}
\end{figure}

With the ever growing need to accommodate a large number of downstream tasks it has become imperative to deploy an LLM with a large number of task-specific adapters.
Several adapters have been proposed in the literature within the framework of parameter efficient fine-tuning (PEFT) \cite{houlsby2019peft, mangrulkar2022peft} such as prefix or prompt tuning, serial adapters, parallel adapters, low-rank adapters (LoRA)~\cite{hu-etal-2023-llm}.
Out of these LoRA has been one of the most widely used adapters for LLM fine-tuning.
These task-specific adapters often constitute a small percentage (less than 1-2\%) of the base model parameter count.
However, these apparently insignificant number of adapter computations introduce a disproportionately significant latency overhead during inference.
Also, it is to be noted that these task specific adapters cannot be merged into the base model a priori, nor can they be merged and unmerged on-the-fly dynamically without incurring significant latency overheads.

In order to highlight the significance of this problem, LLM inference latencies namely time-to-first-token (TTFT) (or prefix-latency or first-token latency) and time-per-output-token (TPOT) (or decode-latency or per-token latency) for 3 different model sizes (1B, 3B and 8B from the LLaMA family) when using the popular LoRA adapters are shown in Fig.~\ref{fig:vllmlat}, as a percentage of the base model latencies.
The latencies are measured using the vLLM inference engine~\cite{kwon2023efficient} at FP16 precision on an NVIDIA H100 GPU, when adapters are attached to all linear projection layers of the base model.
It can be seen that LoRA adapters incur first-token prefill latencies as large as 1.3-2.5x times that of the base model, and per-token decode latencies from 1.3-1.6x times the base model.
More details of this latency measurement experiment are discussed in Sec.~\ref{sec:vllmlat}.
The actual latency measurements (in ms) and the corresponding plots for all models and context lengths are given in Appendix~\ref{app:vllm}.
In order to reduce this large latency overheads it is a common practice to reduce the number of adapter modules by optimizing the placement of adapters such as attaching adapters only to selected transformer layers and to selected linear projection layers (only MHA, only FFN, only QV projection layers, etc) within a transformer layer, often at the expense of accuracies especially for complex tasks.
In view of this, we propose a new zero-latency fused low-rank adapter (zFLoRA) that introduces zero or negligible latency overhead as can be seen in Fig.~\ref{fig:vllmlat}.

The main idea in zFLoRA is to fuse the adapter blocks with the base model projection layers and render the multiplication with input hidden embeddings as a single matmul operation instead of two separate matmuls.
This utilizes the fact that the GPU/NPU hardware is highly optimized for efficient multiplication of large matrices, and shows negligible increase in the cost of matmul when you increase one of the dimensions of a large matrix by a small amount. 
Simultaneous deployment of base model and adapter matmuls also helps reduce any separate memory ops that may be required to copy the inputs and outputs back and forth from the high bandwidth memory.

This can lead to what can be called as a family of fused low-rank adapters (FLoRA).
However, most naive designs would need an expansion of input or reduction of output dimensions for each adapter layer after each fused matmul operation.
In view of this, the architecture of zFLoRA is carefully designed so as to avoid any seemingly trivial operations such as, reducing output dimension by adding/merging the adapter output to the base model output, or expanding the input, which can otherwise cause significant latency overheads.
More details on zFLoRA will be presented in Sections~\ref{sec:fused} and \ref{sec:zflora}.

\section{Related Work}
\label{sec:relatedwork}
Parameter-efficient fine-tuning (PEFT) methods are widely used to adapt or steer the performance of an LLM towards higher accuracies for a specific task~\cite{houlsby2019peft, mangrulkar2022peft}.
PEFT involves learning a small set of augmented parameters or embeddings using a task specific dataset while keeping the whole or a majority of the base model parameters frozen.

Low-rank adapters (LoRA), currently the most commonly used PEFT method, was first introduced in \citet{hu2022lora} based on the hypothesis that weight updates during a downstream task fine-tuning have a low "intrinsic rank." 
With the great success of LoRA, many derivative works which improve on various aspects of the LoRA have been published. 
A comprehensive summary of LoRA and its variants is provided in the survey paper, \citet{Mao_2024}.

Here, we introduce an inexhaustive list of LoRA variants. 
A set of works modify the training scheme, for example, using different learning rates for $A$ and $B$ matrices~\cite{hayou2024lora+}, adding residual connections during training and merge during inference~\cite{shi2024reslora}, or freezing the $A$ matrix and training only $B$ matrix to reduce the memory footprint of training~\cite{zhang2023lora}.
There are another group of studies which concentrate on the low-rank value optimization, such as dynamical rank allocation utilizing SVD of updates~\cite{zhang2023adalora}, adaptive parameter addition~\cite{zhang2023increlora}, and using gating techniques during training based on importance and only keep the most important ranks in the end~\cite{ding2023sparse}. 
\citet{meng2025pissa} optimizes the initialization of LoRA matrices, using principal components of the original weight matrix to initialize $A$ and $B$ and use the residual weight as the frozen weight.

While these works aim to optimize the LoRA's performance, they all preserve the basic structure of LoRA. 
We instead investigate on modifying the structure of LoRA itself. 
This is because our main motivation is to suggest an efficient adapter which can maximize the parallelization of GPUs. 

Parallel adapters~\cite{hetowards} are modules connected to either or both the multi-head attention (MHA) or feed-forawrd network (FFN) blocks.
As the name suggests, parallel adapters are linked in parallel in the graph, that is, the input is shared with the attention (FFN) block and the output is added to that of the attention (FFN). 
Typically the adapter consists of a feed-forward down projection, nonlinearity, and a feed-forward up projection. 
\citet{hu-etal-2023-llm} thoroughly investigates the parallel adapter and concludes that in optimal settings its performance matches with LoRA of similar parameter budget.

In this paper, we do no rely on a single type of adapter. 
Rather, we build upon the parallel adapters' expressive power and use it to complement LoRA.
First, we modify LoRA with the intension of efficient inference and less latency, with the possibility of performance drop.
Then we minimally apply the parallel adapter to counterbalance the loss in performance.
Details of the overall strategy will follow in the next section. 

PEFT includes other methods such as prefix or prompt-tuning~\cite{li2021prefix, lester2021power, liu2022p}, where task-dependent learnable embeddings are appended at the beginning of the context. 
Series adapters~\cite{houlsby2019parameter, pfeiffer2020mad} serially insert additional trainable modules to the `attention$-$FFN' sequence in a layer. 
Survey papers~\cite{xu2023parameter, balne2024parameter} are available for comprehensive list of PEFT methods. 

\section{Family of fused adapters}
\label{sec:fused}
\begin{figure}[t]
    \centering    
        \includegraphics[width=6cm]{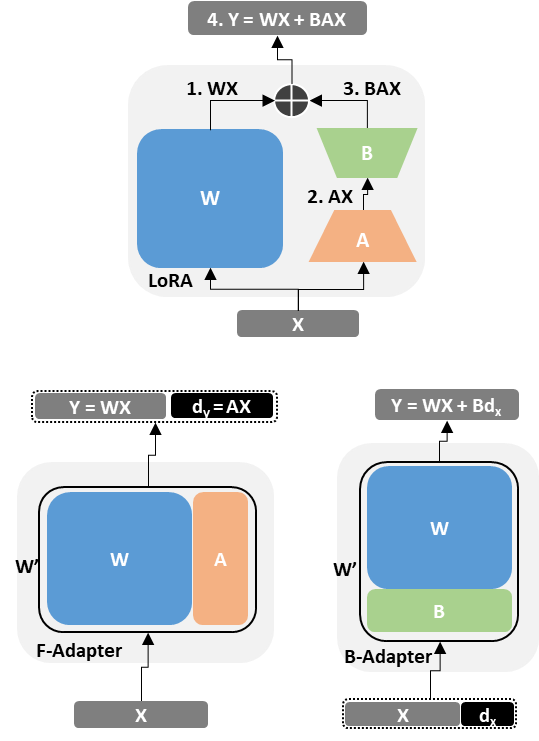}   
    \caption{\small Block schematic of LoRA, and the basic building blocks of a fused adapter (F-Adapter and B-Adapter) for a single projection layer.}
    \label{fig:bblocks}
\end{figure}

Conventional low-rank adapters (LoRA) use low-rank approximation (LRA) in order to process and capture information efficiently in a typically large hidden input dimension using a small number of parameters.
The block schematic of LoRA, and the basic building blocks of a fused adapter namely forward and backward-adapters are shown Fig.~\ref{fig:bblocks}.
For instance, the output of a linear projection layer with weights $W \in {\mathbb R}^{d_o \times d_i}$ and LoRA adapters $A\in {\mathbb R}^{r \times d_i}$, $B\in {\mathbb R}^{d_o \times r}$, for an input $X\in {\mathbb R}^{d_i \times L}$ is given by
\begin{align}
    Z &= WX + BAX
\end{align}
where $d_i$ and $d_o$ are the input and output dimensions, $L$ is the input sequence length, and $r (\ll d_i$ and $d_o$) is the rank of the LRA of the adapter weight matrix $\Delta W = BA$.
The down and up projection matrices $A$ and $B$ may also be referred to as forward and backward adapters, respectively.

\subsection{Partially-fused LoRA}
In a naive implementation of LoRA, the above computation of a single LoRA is performed as a sequence of 4 different operations, namely, $WX$, $AX$, $B(AX)$, and $WX + BAX$.
It is often seen that the overall latency incurred in executing these sequences of operations separately is much larger compared to the total FLOPs that need to be computed.
In order to reduce the overall latency of this compute, and utilize the efficiency of GPUs in parallelization of large size matrix multiplications, the first two operations can be fused into one by concatenating the weight matrices W and A into one.
The resulting computations are given by
\begin{align}
\begin{bmatrix} Y \\ \Delta Y \end{bmatrix}
= \begin{bmatrix} W \\ A \end{bmatrix} X
= \begin{bmatrix} WX \\ AX \end{bmatrix}
\label{eq:pflora}
\end{align}
where $Y=WX$ and $\Delta Y = AX$.
However, the other two operations $\Delta Z = B\Delta Y$ and $Z = Y+\Delta Z$ still need to be computed sequentially.
We refer this way of implementing LoRA as partially-fused LoRA (pf-LoRA).

\begin{figure}[t]
    \centering    
        \includegraphics[width=6.5cm]{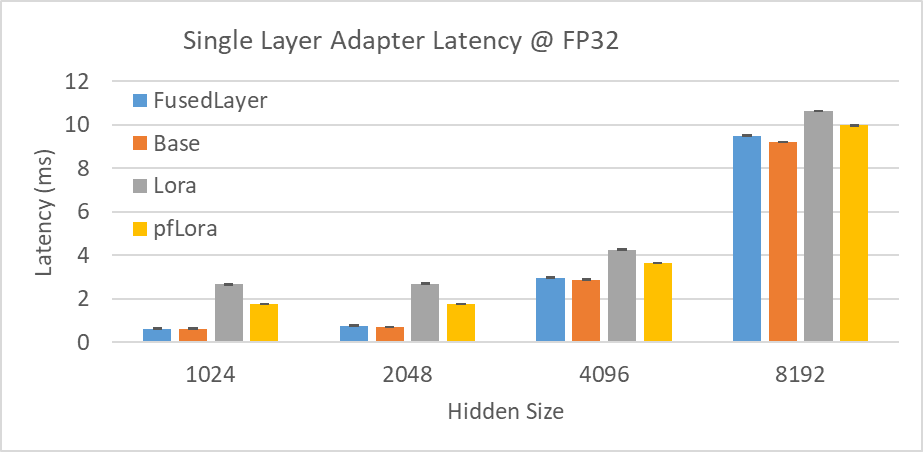} \\
        \includegraphics[width=6.5cm]{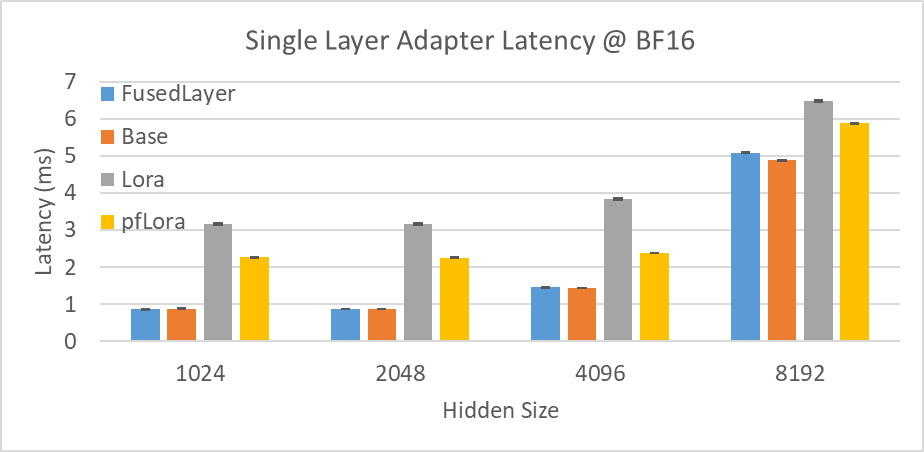}
    \caption{\small Single layer adapter latency simulations for base model layer, LoRA, pfLoRA and a fused layer.}
    \label{fig:singlelayerlat}    
\end{figure}

In order to illustrate the effect of fusing on latency, a single layer simulation of the base layer projection, vanilla LoRA, pf-LoRA, and a fused-adapter layer without any input expansion or output merge operation is conducted.
A single layer forward pass is simulated 100 times equivalent to decoding 100 tokens, and this is iterated 100 times equivalent to processing 100 requests.
The 95 percentile mean latency of this single layer simulation is shown in Fig.~\ref{fig:singlelayerlat}.
It can be seen that both LoRA and pf-LoRA have significant overhead compared to the base layer latencies, while the fused-adapter simulation shows almost negligible overhead.
The fused-adapter simulation is where the base model layer is fused with either the up or down adapter projection as shown in Fig.~\ref{fig:bblocks}.

\subsection{Fused forward adapters}
One way of further reducing the overall latency is to eliminate the LRA framework and remove the backward projection, $B$.
The saved parameter count can be added to the forward projection matrix $A$ by increasing the low-rank dimension from $r$ to $2 r$.
This may be referred to as fused forward adapter (FFA).
In this case, after calculating Eq.~\ref{eq:pflora} we would need one additional computation $Z = Y + Repeat(\Delta Y)$ in order to combine the concatenated outputs obtained from base model ($Y$) and adapter ($\Delta Y$).
The specific operation used to reduce the $d+r$ output to $d$ dimensions can be a design choice, and one option is to repeat the $\Delta Y$ vector $d/2r$ times to match the dimensions of the two vectors and add them.

While FFA can reduce the overall latency, it still has two limitations.
One, without the LRA bottleneck the ability of the adapter module to effectively capture the additional information may reduce significantly during fine-tuning.
The other issue is that, the output of FFA is of dimension $d+r$ and needs to be reduced to $d$ dimensions by merging (repeat and add) the adapter component to the base model component.
This merging operation can introduce non-trivial additional latencies similar to pf-LoRA.

\subsection{Fused backward adapters}
Similar to FFA, we can also design a fused-backward adapter (FBA), where only the backward adapters ($B$) are attached or fused to any projection layer of the base model.
In this case, we do not need the merge operations at the output as required by FFA, but we need an expand operation at the input to convert a $d$ dimensional input to a $d+r$ dimensional input.
One option for this could be split and merge where we divide the $d$ dimensional input into chunks of dimension $r$, and then average these chunks to generate an $r$ dimensional extension for the input.
As in the case of FFA, FFB has similar limitations namely the lack of a LRA bottleneck and the input expansion introducing additional latencies.

\subsection{Fused forward-backward adapters}
Several different combinations of forward and backward adapters attached to different layers within the transformer layer (attention block or the feedforward block) can be explored.
For instance, forward adapters attached to the QKV projection layers and the backward adapter attached to the output projection within the attention block.
The additional $r$ dimensional output from a forward-adapter layer can be passed on to a subsequent backward-adapter layer by appending to its input.
However, the overhead of reducing the output dimension of a forward adapter layer still persists, without which the rotary positional embedding (RoPE) will have to be expanded to $d+r$ dimensions, negatively affecting the information flow previously learned by the base model.
A fused forward-backward adapter (FFBA) with both forward and backward adapters attached to every base model layer can also be designed.
This can add more parameters to a single layer at negligible compute cost and hence can potentially perform better than FFA or FBA, but the latency overheads will be even more severe as it would need both an input expansion as well as an output merge operation.

\section{Zero-latency fused low-rank adapters}
\label{sec:zflora}
\begin{figure}[t]
    \centering    
        \includegraphics[width=5cm]{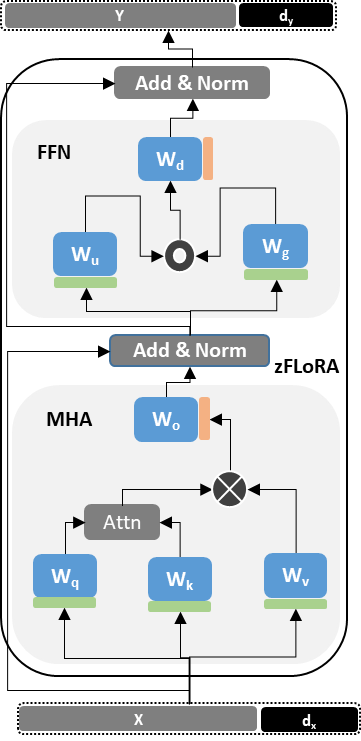}
    \caption{\small Block schematic of zFLoRA architecture within a single transformer block or layer.}
    \label{fig:zflora}    
\end{figure}

In view of the issues associated with naively designed fused adapters outlined above, we propose a carefully designed fused-adapter architecture which retains the forward and backward low-rank approximation, while at the same time eliminates the need for expanding the inputs of a backward adapter layer or reducing the output dimensions of a forward adapter layer.
The block schematic of the proposed zero-latency low-rank adapter (zFLoRA) within a single transformer block or layer is shown in Fig.~\ref{fig:zflora}.

In a naive design of fused forward-backward adapters, one is inclined to attach the forward adapters to the earlier layers such as the QKV projection layers, and the corresponding backward adapter to the output projection layer.
Similarly, forward adapters would be attached to the down and gate projection layers while the backward adapter is attached to the up projection.
As discussed in the previous section, this would need an expansion of input to the QKV projections and merging of output of these forward adapter layers, especially in the attention block, so as to not affect the RoPE embeddings computations.

In order to avoid these seemingly trivial operations that can cause significant latency overheads, we propose to attach the backward adapters first and the forward adapters later within the attention block or the feed-forward block.
This avoids the need for expanding the inputs to QKV projection layers, as the expanded hidden representation from the previous transformer layer (more specifically down-projection of the previous FFN block) is carried forward through layer-norm after the addition of residual component.
Also, since the backward adapter layers yield an automatically merged output there is no need for an additional merge operation for the QKV projections.
However, in this zFLoRA design, the input dimensions need to expanded once before the first transformer layer and needs to be merged back into $d$ dimensions after the last transformer layer before the LM head.
This is a great saving in compute time unlike doing these expand and merge operations for every adapter layer.

In zFLoRA, the pairing of the forward and backward adapters are now spanning across MHA and FFN blocks unlike a naive design which may try to keep them within the MHA or FFN block.
This can also be viewed as a variant of the parallel adapters where the forward and backward adapters are fused with the base projections, the forward-backward pairing is not confined to within a sub-block such as MHA or FFN blocks, without any non-linearity at the LRA bottleneck, and the order of forward and backward adapters apparently inverted within the MHA or FFN block.

\section{Experiments and results}
The performance of the proposed zero-latency fused low-rank adapters is evaluated on 18 different tasks spanning 3 different category of tasks, namely, commonsense reasoning, math reasoning and summary-dialogue generation.
Details of the experimental setup, datasets used, and the results are presented in this section.

\subsection{Datasets}
For commonsense and math reasoning tasks, we use the Commonsense170K and Math10K training datasets used in ~\cite{hu-etal-2023-llm}.
For summary-dialogue tasks we use a combination of training sets from 4 different tasks, namely, CNN-DailyMail, Xsum~\cite{nallapati-etal-2016-abstractive}, DailyDialogue~\cite{li2017dailydialog}, and MultiWoz~\cite{budzianowski2018large}.

\subsection{Experimental setup}
All experiments in this paper are conducted using the publicly available LLaMA family of LLM models~\cite{dubey2024llama, metaLlama32}.
The instruction fine-tuned variants of the models, namely, Llama3.2-1B-Inst and Llama3.2-3B-Inst are used for smaller and latest models.
Adapters were fine-tuned separately for each of the 3 category of tasks on a single node of 8 H100 GPUs with a global batch size of 1M tokens.
All adapters were fine-tuned for 5 epochs for commonsense tasks, 10 epochs for math reasoning tasks, and 3 epochs for the summary and dialogue tasks.
Different learning rates (LR) in the range $1e-6$ to $1e-3$ were explored using a coarse search followed by a fine search for each of the adapters.
A constant LR scheduling with an initial warmup was used for all experiments.
The adapter checkpoints are saved at the end of each epoch and the best performing checkpoint on a heldout validation set is used for final evaluation.
All fine-tuning experiments and evaluations were conducted using our custom implementation of adapters on top of HuggingFace transformers.

\begin{table}[t]
    \centering
    {\small 
    \tabcolsep=0.11cm
    \scalebox{0.9}{
    \begin{tabular}{l|cccccccc|c}
    \hline
     & \multicolumn{8}{c|}{Commonsense Reasoning Tasks (Acc \%) } \\
         Adapter &  arcc & arce & boolq & hella & obqa & piqa & siqa & wino & Avg\\\hline
         \multicolumn{10}{c}{Llama3.2-1B-Inst} \\\hline
         Base & 51.0 & 73.0 & 64.0 & 44.0 & 74.5 & 72.5 & 50.0 & 45.0 & 59.2 \\ 
         FFT & 64.5 & 78.7 & 84.1 & 76.3 & 87.2 & 77.8 & 72.4 & 69.6 & 76.3 \\         
         LoRA & 63.9 & 78.6 & 82.3 & 76.0 & 86.4 & 77.5 & 75.5 & 69.1 & 76.1 \\         
         zFLoRA & 62.8 & 78.4 & 82.6 & 76.9 & 87.4 & 77.3 & 73.1 & 70.1 & 76.1 \\\hline\hline
         \multicolumn{10}{c}{Llama3.2-3B-Inst} \\\hline
         Base & 79.0 & 83.0 & 83.0 & 68.0 & 83.0 & 72.5 & 68.5 & 54.0 & 73.8 \\
         FFT  & 79.0 & 86.4 & 89.3 & 85.4 & 93.2 & 84.7 & 80.4 & 83.2 & 85.2 \\
         LoRA & 77.6 & 86.0 & 89.2 & 84.9 & 93.0 & 85.4 & 80.8 & 84.5 & 85.1 \\
         zFLoRA & 78.2 & 88.2 & 88.1 & 86.1 & 94.0 & 82.7 & 80.7 & 83.6 & 85.2 \\
    \end{tabular}
    }
    }
    \caption{Performance of zFLoRA on commonsense reasoning tasks.}
    \label{tab:cs}
\end{table}
\subsection{Results on 1B and 3B models}
The performance of the proposed zFLoRA on 3 important category of downstream tasks is presented in this section.
The zFLoRA has a strong similarity with LoRA and parallel adapters, and it was shown in ~\cite{hu-etal-2023-llm} that these two adapters performed best as compared to serial adapter and prefix tuning methods. 
In view of this, we provide a comparison of zFLoRA against the base model, full fine-tuning (FFT) and the widely used LoRA.
The primary objective of these experiments is to demonstrate that the proposed zFLoRA performs as close to FFT as possible, and at least as good as LoRA (or parallel adapters) without the latency overheads.

{\bf Commonsense reasoning} is one of the easiest and widely used multiple-choice question-and-answering (Q\&A) tasks used to evaluate the performance of LLMs.
The performance of different adapters for the Llama3.2-1B-Inst and Llama3.2-3B-Inst models on the popular commonsense reasoning tasks when fine-tuned using different adapters is given in Table~\ref{tab:cs}.
As can be seen from the results, full fine-tuning (FFT) of the models perform the best as compared to fine-tuning using adapters.
Barring some minor fluctuations within each task, the proposed zFLoRA performs almost similar to full fine-tuning as well as LoRA.

\begin{table}[t]
    \centering
    {\small
    \tabcolsep=0.11cm
    \scalebox{0.93}{
    \begin{tabular}{l|cccccc|c}
    \hline
     & \multicolumn{6}{c|}{Math Reasoning Tasks (Acc \%) } \\
         Adapter &  addsub & aqua & arith & gsm8k & singeq & svamp & Avg\\\hline
         \multicolumn{8}{c}{Llama3.2-1B-Inst} \\\hline
         Base & 68.10 & 22.83 & 62.17 & 45.49 & 80.91 & 53.20 & 55.45 \\
         FFT  & 85.32 & 22.83 & 96.17 & 48.52 & 90.94 & 66.70 & \bf 68.41 \\
         LoRA & 82.78 & 28.35 & 92.67 & 48.14 & 87.99 & 67.00 & 67.82 \\         
         zFLoRA & 87.85 & 24.80 & 96.00 & 43.37 & 91.93 & 59.40 & 67.22 \\
         \hline\hline
         \multicolumn{8}{c}{Llama3.2-3B-Inst} \\\hline
         Base & 91.14 & 24.80 & 93.17 & 76.88 & 93.90 & 87.60 & \bf 77.91 \\
         FFT  & 89.62 & 28.74 & 99.00 & 71.87 & 93.70 & 82.00 & 77.48 \\
         LoRA & 93.16 & 27.17 & 96.67 & 67.10 & 95.87 & 82.50 & 77.07 \\         
         zFLoRA & 90.38 & 29.53 & 97.17 & 70.74 & 93.70 & 81.90 & 77.23 \\
    \end{tabular}
    }}
    \caption{Performance of zFLoRA on math reasoning tasks.}
    \label{tab:arith}
\end{table}
{\bf Math reasoning} tasks are considered a bit more complicated compared to commonsense tasks, and the LLM is often required to generate multiple tokens giving a numerical answer, and in some cases (gsm8k) a chain of thought reasoning used to arrive at the answer.
The performance of the adapters for the two Llama3.2 models on math reasoning tasks is given in Table~\ref{tab:arith}.
A similar trend as was seen in the case of commonsense reasoning evaluations can be seen.
The proposed zFLoRA performs similar to LoRA and both the adapter methods perform inferior but closer to FFT.

It can be seen that the Llama3.2-3B-Inst base model performance for some math reasoning tasks such as gsm8k and svamp are already the best and none of the adapters including full-finetuning can improve upon the base model.
One possibility is that the instruction fine-tuned model is likely to be trained with several math reasoning instruction data, and the Math10K fine-tuning training set used in this paper is not adding any additional diversity or information.
However, the smaller 1B model shows improvement on all tasks.
Using a more complex math reasoning dataset or using LLM model checkpoints that are saved just after pretraining and without any instruction-finetuning can show better improvement as can be seen in the later scaling-up experiments with LLaMA 7B model.

{\bf Summary and dialogue generation} is an important and more complex downstream application of LLMs.
The performance of various adapters on this category of tasks is shown in Table~\ref{tab:summdia}.
It can be seen from the results that the proposed zFLoRA performs simialr to LoRA, while FFT performs the best.
\begin{table}[t]
    \centering
    {\small
    \tabcolsep=0.11cm
    \scalebox{1.0}{
    \begin{tabular}{l|p{7ex}p{7ex}p{7ex}p{7ex}|c}
    \hline
     & \multicolumn{4}{c|}{Summary/Dialogue Tasks ($R_{Lsum}$) } & \\
         Adapter &  cnndm & dd & woz & xsum & Avg\\\hline
         \multicolumn{6}{c}{Llama3.2-1B-Inst} \\\hline
         Base    &  25.28 & 13.03 & 13.81 & 19.49 & 17.90 \\
         FFT    &  28.37 & 16.58 & 30.45 & 32.67 & 27.01  \\
         LoRA    & 26.76 & 20.12 & 31.34 & 32.23	& \bf 27.61 \\         
         zFLoRA    & 27.25 & 18.31 & 31.82 & 30.98 & 27.09 \\\hline\hline
         \multicolumn{6}{c}{Llama3.2-3B-Inst} \\\hline
         Base & 25.10 & 14.45 &  16.68 & 20.54 & 19.19 \\
         FFT & 29.23 & 25.85 &  29.66 & 37.63 & \bf 30.59 \\
         Lora & 28.92 & 18.37 & 31.15 & 36.45 & 28.72 \\         
         zFLoRA & 28.83 & 19.44 & 30.76 & 36.18 & 28.80 \\\hline
    \end{tabular}
    }}
    \caption{Performance of zFLoRA on summary/dialogue tasks.}
    \label{tab:summdia}
\end{table}

{\bf Performance vs rank:} Experimental results on the performance of zFLoRA as against LoRA for 1B and 3B models for varying adapter ranks is given in Appendix~\ref{app:perfvsrank}.

{\bf Performance of FFA and FFBA adapters} which belong to the family of fused adapters or fused low-rank adapters (FLoRA) as compared to the zFLoRA is discussed in Appendix~\ref{app:ffaffba}.

\subsection{Scaling up and comparison experiments}
In order to verify that the proposed zFLoRA adapter scales up to larger LLMs, and to compare its performance against other popular PEFT adapters we conduct experiments using the LLaMA 7B model \cite{touvron2023llamaopenefficientfoundation} with exactly same code and experimental setup as outlined in ~\cite{hu-etal-2023-llm}.
Performance of zFLoRA on the 7B model as compared to other PEFT adaptation methods is shown in Tables~\ref{tab:llama7b-cs} and \ref{tab:llama7b-math}.
The results marked $^+$ are directly reported from \cite{hu-etal-2023-llm}, while the bottom two rows are experiments repeated for LoRA and zFLoRA using the same code and the exact experimental setup (3 epochs and LR 3e-4) used by the authors.
The Base$^*$ results are reported as is from the original LLaMA paper~\cite{touvron2023llamaopenefficientfoundation}.
It can be seen that the repeat LoRA results closely match the results reported in ~\cite{hu-etal-2023-llm}, and our proposed zFLoRA matches the performance of LoRA and parallel adapters quite closely.

\begin{table}[t]
    \centering
    {\small 
    \tabcolsep=0.11cm
    \scalebox{0.9}{
    \begin{tabular}{l|cccccccc|c}
    \hline
        & \multicolumn{8}{c|}{Commonsense Reasoning Tasks (Acc \%) } \\
        Adapter &  boolq & piqa & siqa & hella & wino & arce & arcc & obqa & Avg\\\hline
        Base$^*$ &  76.5 & 79.8 & 48.9 & 76.1 & 70.1 & 72.8 & 47.6 & 57.2 & 66.1 \\
        Prefix$^+$ & 64.3 & 76.8 & 73.9 & 42.1 & 72.1 & 72.9 & 54.0 & 60.6 & 64.6 \\
        Series$^+$ & 63.0 & 79.2 & 76.3 & 67.9 & 75.7 & 74.5 & 57.1 & 72.4 & 70.8 \\
        Parallel$^+$ & 67.9 & 76.4 & 78.8 & 69.8 & 78.9 & 73.7 & 57.3 & 75.2 & 72.3 \\
        LoRA$^+$ & 68.9 & 80.7 & 77.4 & 78.1 & 78.8 & 77.8 & 61.3 & 74.8 & 74.7 \\\hline
        LoRA & 68.4 & 80.8 & 79.1 & 82.5 & 80.0 & 76.9 & 62.0 & 78.2 & {\bf 76.0}\\
        zFLoRA & 69.8 & 78.0 & 79.2 & 79.8 & 81.7 & 78.7 & 62.2 & 78.0 & 75.9 \\
    \end{tabular}
    }
    }
    \caption{Performance of zFLoRA on commonsense reasoning tasks for LLaMA-7B model. $^*$~\cite{touvron2023llamaopenefficientfoundation},  $^+$~\cite{hu-etal-2023-llm}.}
    \label{tab:llama7b-cs}
\end{table}

\begin{table}[t]
    \centering    
    {\small
    \tabcolsep=0.11cm
    \scalebox{0.95}{
    \begin{tabular}{l|cccccc|c}
        \hline
        & \multicolumn{6}{c|}{Math Reasoning Tasks (Acc \%) } \\
        Adapter &  arith & gsm8k & addsub & aqua & singeq & svamp & Avg\\\hline
        Base$^*$ & - & 11.0 & - & - & - & - & - \\
        Prefix$^+$ & 63.2 & 24.4 & 57.0 & 14.2 & 55.3 & 38.1 &  42.0 \\
        Series$^+$ & 92.8 & 33.3 & 80.0 & 15.0 & 83.5 & 52.3 & 59.5 \\
        Parallel$^+$ & 94.5 & 35.3 & 86.6 & 18.1 & 86.0 & 49.6 & 61.7 \\
        LoRA$^+$ & 95.0 & 37.5 & 83.3 & 18.9 & 84.4 & 52.1 & 61.9 \\\hline
        LoRA & 96.2 & 39.7 & 81.0 & 16.9 & 84.1 & 47.3 & 60.9 \\
        zFLoRA & 94.3 & 38.0 & 85.8 & 19.3 & 87.4 & 47.7 & {\bf 62.1} \\
        \end{tabular}
    }}
    \caption{Performance of zFLoRA on math reasoning tasks for LLaMA-7B model. $^*$~\cite{touvron2023llamaopenefficientfoundation}, $^+$~\cite{hu-etal-2023-llm}.}
    \label{tab:llama7b-math}  
\end{table}

\section{Latency measurements}
\label{sec:lat}
\vspace{-2mm}
A comparison and discussion on the inference time latencies of the proposed zFLoRA as compared to the base model and the popular LoRA adapters is provided in this section.
The latency measurements are performed on two different platforms namely, an NVIDIA H100 GPU and a Samsung Galaxy S25+ mobile NPU.

\subsection{Latencies on H100 GPU}
\label{sec:vllmlat}
The inference latencies were measured using the vLLM inference engine popularly used to deploy small to medium sized commercial LLMs on different GPU and edge platforms~\cite{kwon2023efficient}.
The time-to-first-token (TTFT) and time-per-output-token (TPOT) latencies are measured for models of different size (1B, 3B and 8B) from the LLaMA-3.x family.
The latencies are measured on an NVIDIA H100 GPU with 80GB memory using vLLM's online serving mode.
Latencies are measured by passing 100 random input prompts of fixed length to the inference engine to generate 128 output tokens, with a maximum concurrency of 1 (batch size 1).
Experiments were repeated for different input lengths ranging from 512 to 8192.
Latencies were measured for the base models without any adapters, and with adapters LoRA and zFLoRA separately.
An adapter rank of 32 was used and the adapters were applied to all linear layers within a transformer block.
The resulting number of parameters for LoRA/zFLoRA were 22.5M/15M (2.25\%/1.5\%), 48.6M/29.4M (1.6\%/0.98\%), 83.9M/54.5M (1.04\%/0.68\%) for the 1B, 3B and 8B models, respectively.
The measured latencies are shown in Fig.~\ref{fig:vllmlat} relative to the base model latencies as a percentage.
It can be clearly seen that zFLoRA has almost zero to negligible latency overhead and decodes almost at the same speed as the base model, while LoRA introduces significant overheads as discussed in Section~\ref{sec:intro}.
The actual latencies measured (in ms) and the corresponding plots are shown in Appendix~\ref{app:vllm}.

\begin{figure}[t]
    \centering    
    \begin{tabular}{cc}
        \includegraphics[width=3.8cm]{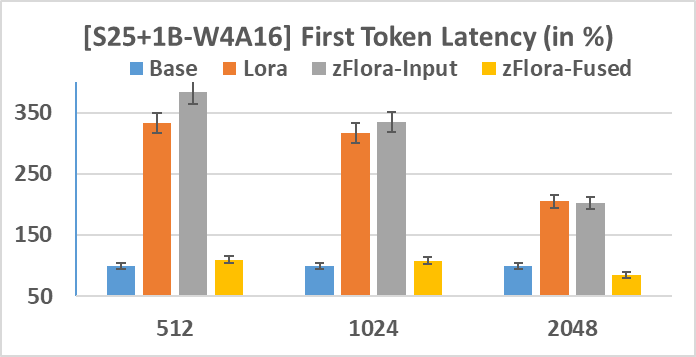} &
        \includegraphics[width=3.8cm]{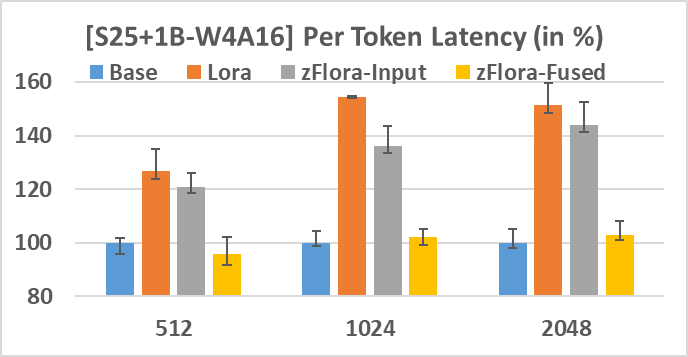} \\
        \includegraphics[width=3.8cm]{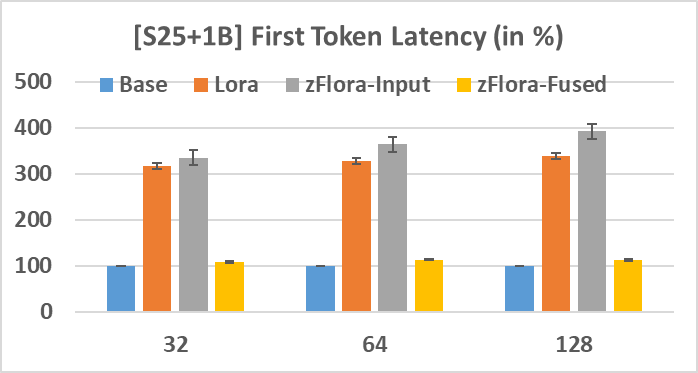} &
        \includegraphics[width=3.8cm]{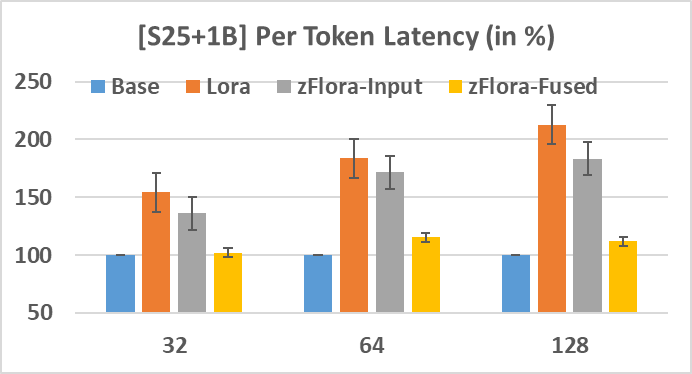} 
    \end{tabular}
    \vspace{-2mm}
    \caption{\small On-device prefill and decode latencies of LoRA and zFLoRA for varying prompt lengths (top row) and adapter ranks (bottom row), as compared to the base model (1B) on Samsung Galaxy S25+ mobile handset.}
    \label{fig:s25lat}    
    \vspace{-5mm}
\end{figure} 
\subsection{Latencies on Samsung Galaxy S25+ NPU}
\label{sec:npulat}
The inference graphs for the base model, as well as LoRA and zFLoRA adapters are frozen with a 4-bit quantization for the base model weights and an activation quantization of 16-bits.
The S25+ (Qualcomm Snapdragon 8 Elite NPU) latencies of adapters for varying context lengths (512 to 2048) and ranks (32 to 128) as compared to the base model is shown in Fig.~\ref{fig:s25lat}.
The frozen graph is used to decode 10 random prompts with varying context lengths and generating 10 tokens per prompt.
A fixed context-length of 1024 is used for latency measurements with varying adapter ranks.
Owing to the current limitations of the Qualcomm APIs which do not support efficient and dynamic loading or swapping of weights, adapter weights are passed as 16-bit inputs to the graph along with the prompt embeddings.
In view of this, it can be seen that both LoRA and zFLoRA-Input show significant latency overheads compared to the base model.
Latest Qualcomm APIs support a new feature for dynamic (or partial) loading of only the adapter weights in a frozen graph, however, this feature is still not fully optimized.
We hope this feature to be more optimized in the future and/or Qualcomm provides options for partial replacement of frozen weights or dynamic concatenation of weights at runtime, that will enable realizing the zero-latency potential of zFLoRA-Fused as shown in the figure.
Latencies for zFLoRA-Fused are measured by quantizing both the model and adapter weights to 4-bits and the activation to 16-bits.
Detailed measurements of the latencies (in ms) for both 1B and 3B models is given in Appendix~\ref{app:s25}.

\section{Conclusions}
In this paper, we proposed a novel zero-latency fused low-rank adapter (zFLoRa) for fine-tuning LLMs to downstream tasks.
The proposed zFLoRA adapters can be viewed as a combination of ideas from fused matmul operations, low-rank approximation, block-level parallel adapters, layer-level LoRA style adapters, and also involves careful design or placement of forward and backward adapter components so as to eliminate any merge or expand operations on the input or output embeddings.
Experimental results and latency measurements (on GPU as well as NPU) using models from 1B to 7B show that zFLoRA matches the performance of the widely used LoRA, while having zero-latency overhead at inference time.
Several variants of the proposed zFLoRA can be explored to further reduce the overall adapter parameter count.
Some obvious choices are using adapters only on MHA blocks, and on only selected layers (first, last, mid or alternative).
The proposed zFLoRA solution can be deployed as it is on GPU or edge platforms for zero-latency overhead, however on-device deployment on NPU platforms would need additional support from NPU developers for partial replacement of weights in a frozen graph or dynamic loading and concatenation of adapter weights to the base model weights.

\section{Limitations}
We recognize the following limitations of our work.
The experiments and down-stream applications considered in this paper are restricted to one language (English), one modality (text) and can be extended to other languages and modalities.
The zFLoRA method may be more relevant to small or moderately sized LLMs (1B to 7B parameters) that could be candidates for on-device deployment and with single prompt/task decoding (batch size 1).
ZFLoRA can be applied for batch decoding over a homogeneous set of tasks using the same adapter modules, however it cannot be applied to a heterogeneous set of tasks.
Experiments with huge cloud based LLMs and larger batch size (serving the same task) is possible, but the significance of latency overheads and need for optimization has to be investigated carefully, which is out of scope of this paper.
In this paper, we compare vanilla-zFLoRA with vanilla-LoRA for performance.
However, more recent studies such as LoRA-Pro \cite{wang2025loraprolowrankadaptersproperly} claim to bridge the gap between vanilla-LoRA and FFT, albeit with older generation models such as LLaMA-2.
A more detailed comparison of zFLoRA with LoRA-Pro using latest models and datasets, and the possibility of extending LoRA-Pro and similar refinements to zFLoRA are part of future study.
The multi-adapter zFLoRA solution can be readily deployed on GPU/CPU based edge solutions, but has some limitations on NPU platforms.
See Sec.~\ref{sec:npulat} for more details.
We do hope the potential latency benefits will motivate the NPU hardware/compiler developers to support dynamic fusing of base and adapter weights in their future releases.

\bibliography{custom, mybib}

\begin{thebibliography}{34}
\providecommand{\natexlab}[1]{#1}

\bibitem[{Allal et~al.(2025)Allal, Lozhkov, Bakouch, Blázquez, Penedo, Tunstall, Marafioti, Kydlíček, Lajarín, Srivastav, Lochner, Fahlgren, Nguyen, Fourrier, Burtenshaw, Larcher, Zhao, Zakka, Morlon, Raffel, von Werra, and Wolf}]{allal2025smollm2smolgoesbig}
Loubna~Ben Allal, Anton Lozhkov, Elie Bakouch, Gabriel~Martín Blázquez, Guilherme Penedo, Lewis Tunstall, Andrés Marafioti, Hynek Kydlíček, Agustín~Piqueres Lajarín, Vaibhav Srivastav, Joshua Lochner, Caleb Fahlgren, Xuan-Son Nguyen, Clémentine Fourrier, Ben Burtenshaw, Hugo Larcher, Haojun Zhao, Cyril Zakka, Mathieu Morlon, Colin Raffel, Leandro von Werra, and Thomas Wolf. 2025.
\newblock \href {https://arxiv.org/abs/2502.02737} {Smollm2: When smol goes big -- data-centric training of a small language model}.
\newblock \emph{Preprint}, arXiv:2502.02737.

\bibitem[{Balne et~al.(2024)Balne, Bhaduri, Roy, Jain, and Chadha}]{balne2024parameter}
Charith Chandra~Sai Balne, Sreyoshi Bhaduri, Tamoghna Roy, Vinija Jain, and Aman Chadha. 2024.
\newblock Parameter efficient fine tuning: A comprehensive analysis across applications.
\newblock \emph{arXiv preprint arXiv:2404.13506}.

\bibitem[{Budzianowski et~al.(2018)Budzianowski, Wen, Tseng, Casanueva, Stefan, Osman, and Ga{\v{s}}i\'c}]{budzianowski2018large}
Pawe{\l} Budzianowski, Tsung-Hsien Wen, Bo-Hsiang Tseng, I{\~n}igo Casanueva, Ultes Stefan, Ramadan Osman, and Milica Ga{\v{s}}i\'c. 2018.
\newblock Multiwoz - a large-scale multi-domain wizard-of-oz dataset for task-oriented dialogue modelling.
\newblock In \emph{Proceedings of the 2018 Conference on Empirical Methods in Natural Language Processing (EMNLP)}.

\bibitem[{DeepSeek-AI et~al.(2025)DeepSeek-AI, Liu, Feng, Xue, Wang, Wu, Lu, Zhao, Deng, Zhang, Ruan, Dai, Guo, Yang, Chen, Ji, Li, Lin, Dai, Luo, Hao, ..., and Pan}]{deepseekai2025deepseekv3technicalreport_etal}
DeepSeek-AI, Aixin Liu, Bei Feng, Bing Xue, Bingxuan Wang, Bochao Wu, Chengda Lu, Chenggang Zhao, Chengqi Deng, Chenyu Zhang, Chong Ruan, Damai Dai, Daya Guo, Dejian Yang, Deli Chen, Dongjie Ji, Erhang Li, Fangyun Lin, Fucong Dai, Fuli Luo, Guangbo Hao, ..., and Zizheng Pan. 2025.
\newblock \href {https://arxiv.org/abs/2412.19437} {Deepseek-v3 technical report}.
\newblock \emph{Preprint}, arXiv:2412.19437.

\bibitem[{Ding et~al.(2023)Ding, Lv, Wang, Chen, Zhou, Liu, and Sun}]{ding2023sparse}
Ning Ding, Xingtai Lv, Qiaosen Wang, Yulin Chen, Bowen Zhou, Zhiyuan Liu, and Maosong Sun. 2023.
\newblock Sparse low-rank adaptation of pre-trained language models.
\newblock \emph{arXiv preprint arXiv:2311.11696}.

\bibitem[{Gemma-Team et~al.(2025)Gemma-Team, Kamath, and et~al.}]{gemmateam2025gemma3technicalreport}
Gemma-Team, Aishwarya Kamath, and et~al. 2025.
\newblock \href {https://arxiv.org/abs/2503.19786} {Gemma 3 technical report}.
\newblock \emph{Preprint}, arXiv:2503.19786.

\bibitem[{Grattafiori et~al.(2024)Grattafiori, Dubey, Jauhri, Pandey, Kadian, Al-Dahle, Letman, Mathur, Schelten, Vaughan, Yang, Fan, Goyal, Hartshorn, Yang, Mitra, Sravankumar, Korenev, Hinsvark, and other authors}]{dubey2024llama}
Aaron Grattafiori, Abhimanyu Dubey, Abhinav Jauhri, Abhinav Pandey, Abhishek Kadian, Ahmad Al-Dahle, Aiesha Letman, Akhil Mathur, Alan Schelten, Alex Vaughan, Amy Yang, Angela Fan, Anirudh Goyal, Anthony Hartshorn, Aobo Yang, Archi Mitra, Archie Sravankumar, Artem Korenev, Arthur Hinsvark, and 400+ other authors. 2024.
\newblock The llama 3 herd of models.
\newblock \emph{https://arxiv.org/abs/2407.21783}.

\bibitem[{Hayou et~al.(2024)Hayou, Ghosh, and Yu}]{hayou2024lora+}
Soufiane Hayou, Nikhil Ghosh, and Bin Yu. 2024.
\newblock Lora+: Efficient low rank adaptation of large models.
\newblock \emph{arXiv preprint arXiv:2402.12354}.

\bibitem[{He et~al.(2022)He, Zhou, Ma, Berg-Kirkpatrick, and Neubig}]{hetowards}
Junxian He, Chunting Zhou, Xuezhe Ma, Taylor Berg-Kirkpatrick, and Graham Neubig. 2022.
\newblock Towards a unified view of parameter-efficient transfer learning.
\newblock In \emph{International Conference on Learning Representations}.

\bibitem[{Houlsby et~al.(2019{\natexlab{a}})Houlsby, Giurgiu, Jastrzebski, Morrone, de~Laroussilhe, Gesmundo, Attariyan, and Gelly}]{houlsby2019peft}
Neil Houlsby, Andrei Giurgiu, Stanislaw Jastrzebski, Bruna Morrone, Quentin de~Laroussilhe, Andrea Gesmundo, Mona Attariyan, and Sylvain Gelly. 2019{\natexlab{a}}.
\newblock \href {https://arxiv.org/abs/1902.00751} {Parameter-efficient transfer learning for nlp}.
\newblock \emph{Preprint}, arXiv:1902.00751.

\bibitem[{Houlsby et~al.(2019{\natexlab{b}})Houlsby, Giurgiu, Jastrzebski, Morrone, De~Laroussilhe, Gesmundo, Attariyan, and Gelly}]{houlsby2019parameter}
Neil Houlsby, Andrei Giurgiu, Stanislaw Jastrzebski, Bruna Morrone, Quentin De~Laroussilhe, Andrea Gesmundo, Mona Attariyan, and Sylvain Gelly. 2019{\natexlab{b}}.
\newblock Parameter-efficient transfer learning for nlp.
\newblock In \emph{International conference on machine learning}, pages 2790--2799. PMLR.

\bibitem[{Hu et~al.(2022)Hu, Shen, Wallis, Allen-Zhu, Li, Wang, Wang, and Chen}]{hu2022lora}
Edward~J Hu, Yelong Shen, Phillip Wallis, Zeyuan Allen-Zhu, Yuanzhi Li, Shean Wang, Lu~Wang, and Weizhu Chen. 2022.
\newblock \href {https://openreview.net/forum?id=nZeVKeeFYf9} {Lo{RA}: Low-rank adaptation of large language models}.
\newblock In \emph{International Conference on Learning Representations}.

\bibitem[{Hu et~al.(2023)Hu, Wang, Lan, Xu, Lim, Bing, Xu, Poria, and Lee}]{hu-etal-2023-llm}
Zhiqiang Hu, Lei Wang, Yihuai Lan, Wanyu Xu, Ee-Peng Lim, Lidong Bing, Xing Xu, Soujanya Poria, and Roy Lee. 2023.
\newblock \href {https://doi.org/10.18653/v1/2023.emnlp-main.319} {{LLM}-adapters: An adapter family for parameter-efficient fine-tuning of large language models}.
\newblock In \emph{Proceedings of the 2023 Conference on Empirical Methods in Natural Language Processing}, pages 5254--5276, Singapore. Association for Computational Linguistics.

\bibitem[{Kwon et~al.(2023)Kwon, Li, Zhuang, Sheng, Zheng, Yu, Gonzalez, Zhang, and Stoica}]{kwon2023efficient}
Woosuk Kwon, Zhuohan Li, Siyuan Zhuang, Ying Sheng, Lianmin Zheng, Cody~Hao Yu, Joseph~E. Gonzalez, Hao Zhang, and Ion Stoica. 2023.
\newblock Efficient memory management for large language model serving with pagedattention.
\newblock In \emph{Proceedings of the ACM SIGOPS 29th Symposium on Operating Systems Principles}.

\bibitem[{Lester et~al.(2021)Lester, Al-Rfou, and Constant}]{lester2021power}
Brian Lester, Rami Al-Rfou, and Noah Constant. 2021.
\newblock The power of scale for parameter-efficient prompt tuning.
\newblock In \emph{Proceedings of the 2021 Conference on Empirical Methods in Natural Language Processing}, pages 3045--3059.

\bibitem[{Li and Liang(2021)}]{li2021prefix}
Xiang~Lisa Li and Percy Liang. 2021.
\newblock Prefix-tuning: Optimizing continuous prompts for generation.
\newblock In \emph{Proceedings of the 59th Annual Meeting of the Association for Computational Linguistics and the 11th International Joint Conference on Natural Language Processing (Volume 1: Long Papers)}, pages 4582--4597.

\bibitem[{Li et~al.(2017)Li, Su, Shen, Li, Cao, and Niu}]{li2017dailydialog}
Yanran Li, Hui Su, Xiaoyu Shen, Wenjie Li, Ziqiang Cao, and Shuzi Niu. 2017.
\newblock Dailydialog: A manually labelled multi-turn dialogue dataset.
\newblock In \emph{Proceedings of The 8th International Joint Conference on Natural Language Processing (IJCNLP 2017)}.

\bibitem[{Liu et~al.(2022)Liu, Ji, Fu, Tam, Du, Yang, and Tang}]{liu2022p}
Xiao Liu, Kaixuan Ji, Yicheng Fu, Weng Tam, Zhengxiao Du, Zhilin Yang, and Jie Tang. 2022.
\newblock P-tuning: Prompt tuning can be comparable to fine-tuning across scales and tasks.
\newblock In \emph{Proceedings of the 60th Annual Meeting of the Association for Computational Linguistics (Volume 2: Short Papers)}, pages 61--68.

\bibitem[{Liu et~al.(2024)Liu, Zhao, Iandola, Lai, Tian, Fedorov, Xiong, Chang, Shi, Krishnamoorthi, Lai, and Chandra}]{liu2024mobilellmoptimizingsubbillionparameter}
Zechun Liu, Changsheng Zhao, Forrest Iandola, Chen Lai, Yuandong Tian, Igor Fedorov, Yunyang Xiong, Ernie Chang, Yangyang Shi, Raghuraman Krishnamoorthi, Liangzhen Lai, and Vikas Chandra. 2024.
\newblock \href {https://arxiv.org/abs/2402.14905} {Mobilellm: Optimizing sub-billion parameter language models for on-device use cases}.
\newblock \emph{Preprint}, arXiv:2402.14905.

\bibitem[{Mangrulkar et~al.(2022)Mangrulkar, Gugger, Debut, Belkada, Paul, and Bossan}]{mangrulkar2022peft}
Sourab Mangrulkar, Sylvain Gugger, Lysandre Debut, Younes Belkada, Sayak Paul, and B~Bossan. 2022.
\newblock Peft: State-of-the-art parameter-efficient fine-tuning methods.
\newblock \emph{URL: https://github. com/huggingface/peft}.

\bibitem[{Mao et~al.(2024)Mao, Ge, Fan, Xu, Mi, Hu, and Gao}]{Mao_2024}
Yuren Mao, Yuhang Ge, Yijiang Fan, Wenyi Xu, Yu~Mi, Zhonghao Hu, and Yunjun Gao. 2024.
\newblock \href {https://doi.org/10.1007/s11704-024-40663-9} {A survey on lora of large language models}.
\newblock \emph{Frontiers of Computer Science}, 19(7).

\bibitem[{Meng et~al.(2025)Meng, Wang, and Zhang}]{meng2025pissa}
Fanxu Meng, Zhaohui Wang, and Muhan Zhang. 2025.
\newblock Pissa: Principal singular values and singular vectors adaptation of large language models.
\newblock \emph{Advances in Neural Information Processing Systems}, 37:121038--121072.

\bibitem[{Meta-AI(2024)}]{metaLlama32}
Meta-AI. 2024.
\newblock {L}lama 3.2: {R}evolutionizing edge {A}{I} and vision with open, customizable models --- ai.meta.com.
\newblock https://ai.meta.com/blog/llama-3-2-connect-2024-vision-edge-mobile-devices/.
\newblock [Accessed 16-02-2025].

\bibitem[{Nallapati et~al.(2016)Nallapati, Zhou, dos Santos, Gu{\ensuremath{\dot{}}}l{\c{c}}ehre, and Xiang}]{nallapati-etal-2016-abstractive}
Ramesh Nallapati, Bowen Zhou, Cicero dos Santos, {\c{C}}a{\u{g}}lar Gu{\ensuremath{\dot{}}}l{\c{c}}ehre, and Bing Xiang. 2016.
\newblock \href {https://doi.org/10.18653/v1/K16-1028} {Abstractive text summarization using sequence-to-sequence {RNN}s and beyond}.
\newblock In \emph{Proceedings of the 20th {SIGNLL} Conference on Computational Natural Language Learning}, pages 280--290, Berlin, Germany. Association for Computational Linguistics.

\bibitem[{OpenAI et~al.(2024)OpenAI, Achiam, and et~al.}]{openai2024gpt4_etal}
OpenAI, Josh Achiam, and et~al. 2024.
\newblock \href {https://arxiv.org/abs/2303.08774} {Gpt-4 technical report}.
\newblock \emph{Preprint}, arXiv:2303.08774.

\bibitem[{Pfeiffer et~al.(2020)Pfeiffer, Vuli{\'c}, Gurevych, and Ruder}]{pfeiffer2020mad}
Jonas Pfeiffer, Ivan Vuli{\'c}, Iryna Gurevych, and Sebastian Ruder. 2020.
\newblock Mad-x: An adapter-based framework for multi-task cross-lingual transfer.
\newblock In \emph{Proceedings of the 2020 Conference on Empirical Methods in Natural Language Processing (EMNLP)}, pages 7654--7673.

\bibitem[{Shi et~al.(2024)Shi, Huang, Song, Li, Zhang, Huang, Wei, Deng, Sun, and Zhang}]{shi2024reslora}
Shuhua Shi, Shaohan Huang, Minghui Song, Zhoujun Li, Zihan Zhang, Haizhen Huang, Furu Wei, Weiwei Deng, Feng Sun, and Qi~Zhang. 2024.
\newblock Reslora: Identity residual mapping in low-rank adaption.
\newblock \emph{arXiv preprint arXiv:2402.18039}.

\bibitem[{Touvron et~al.(2023)Touvron, Lavril, Izacard, Martinet, Lachaux, Lacroix, Rozière, Goyal, Hambro, Azhar, Rodriguez, Joulin, Grave, and Lample}]{touvron2023llamaopenefficientfoundation}
Hugo Touvron, Thibaut Lavril, Gautier Izacard, Xavier Martinet, Marie-Anne Lachaux, Timothée Lacroix, Baptiste Rozière, Naman Goyal, Eric Hambro, Faisal Azhar, Aurelien Rodriguez, Armand Joulin, Edouard Grave, and Guillaume Lample. 2023.
\newblock \href {https://arxiv.org/abs/2302.13971} {Llama: Open and efficient foundation language models}.
\newblock \emph{Preprint}, arXiv:2302.13971.

\bibitem[{Wang et~al.(2025)Wang, Liang, He, Wang, and Tan}]{wang2025loraprolowrankadaptersproperly}
Zhengbo Wang, Jian Liang, Ran He, Zilei Wang, and Tieniu Tan. 2025.
\newblock \href {https://arxiv.org/abs/2407.18242} {Lora-pro: Are low-rank adapters properly optimized?}
\newblock \emph{Preprint}, arXiv:2407.18242.

\bibitem[{Xu et~al.(2024)Xu, Li, Chen, Wang, Gao, Cai, and Ling}]{xu2024ondevicelanguagemodelscomprehensive}
Jiajun Xu, Zhiyuan Li, Wei Chen, Qun Wang, Xin Gao, Qi~Cai, and Ziyuan Ling. 2024.
\newblock \href {https://arxiv.org/abs/2409.00088} {On-device language models: A comprehensive review}.
\newblock \emph{Preprint}, arXiv:2409.00088.

\bibitem[{Xu et~al.(2023)Xu, Xie, Qin, Tao, and Wang}]{xu2023parameter}
Lingling Xu, Haoran Xie, Si-Zhao~Joe Qin, Xiaohui Tao, and Fu~Lee Wang. 2023.
\newblock Parameter-efficient fine-tuning methods for pretrained language models: A critical review and assessment.
\newblock \emph{arXiv preprint arXiv:2312.12148}.

\bibitem[{Zhang et~al.(2023{\natexlab{a}})Zhang, Li, Chen, Jiang, Wang, and Qian}]{zhang2023increlora}
Feiyu Zhang, Liangzhi Li, Junhao Chen, Zhouqiang Jiang, Bowen Wang, and Yiming Qian. 2023{\natexlab{a}}.
\newblock Increlora: Incremental parameter allocation method for parameter-efficient fine-tuning.
\newblock \emph{arXiv preprint arXiv:2308.12043}.

\bibitem[{Zhang et~al.(2023{\natexlab{b}})Zhang, Zhang, Shi, Chu, and Li}]{zhang2023lora}
Longteng Zhang, Lin Zhang, Shaohuai Shi, Xiaowen Chu, and Bo~Li. 2023{\natexlab{b}}.
\newblock Lora-fa: Memory-efficient low-rank adaptation for large language models fine-tuning.
\newblock \emph{arXiv preprint arXiv:2308.03303}.

\bibitem[{Zhang et~al.(2023{\natexlab{c}})Zhang, Chen, Bukharin, Karampatziakis, He, Cheng, Chen, and Zhao}]{zhang2023adalora}
Qingru Zhang, Minshuo Chen, Alexander Bukharin, Nikos Karampatziakis, Pengcheng He, Yu~Cheng, Weizhu Chen, and Tuo Zhao. 2023{\natexlab{c}}.
\newblock Adalora: Adaptive budget allocation for parameter-efficient fine-tuning.
\newblock \emph{arXiv preprint arXiv:2303.10512}.

\end{thebibliography}


\appendix
\begin{table*}[t]
    \centering
    \begin{tabular}{c|c|ccccc|ccccc}
    \hline
        \multicolumn{2}{c|}{} & \multicolumn{5}{c|}{Mean TTFT (ms)} & \multicolumn{5}{c}{Mean TPOT (ms)}\\\hline				
		\multicolumn{2}{c|}{Input len} & 512	& 1024	& 2048	& 4096	& 8192 & 512	& 1024	& 2048	& 4096	& 8192 \\\hline
        \multirow{3}{*}{1B} &	Base & 8.69 & 11.51	& 18.01	& 34.56	& 64.75 & 2.44	& 2.46	& 2.49	& 2.52	& 2.63\\
	& LoRA	& 22.47	& 25.33	& 30.92	& 58.99	& 111.06 & 3.87	& 3.79	& 3.82	& 3.85	& 3.91\\
	& zFLoRA	& 8.8	& 12.06	& 18.58	& 35.07	& 63.79 & 2.45	& 2.46	& 2.47	& 2.53	& 2.62\\\hline
		\multirow{3}{*}{3B} &	Base	& 13.18	& 19.58	& 32.86	& 61.54	& 136.00 & 4.53	& 4.57	& 4.62	& 4.76	& 4.96 \\
	& LoRA	& 34.55	& 36.63	& 50.59	& 95.06	& 201.61 & 6.47	& 6.47	& 6.53	& 6.65	& 6.85\\
	& zFLoRA	& 13.96	& 19.36	& 31.36	& 60.33	& 130.28 & 4.56	& 4.56	& 4.63	& 4.73	& 4.9\\\hline
        \multirow{3}{*}{8B} & Base	& 22.78	& 35.18	& 62.32	& 123.49	& 267.46 & 7.52	& 7.54	& 7.6	& 7.73	& 7.93 \\
	& LoRA	& 37.42	& 50.06	& 87.82	& 170.89	& 353.89 & 10.06	& 10.1	& 10.19	& 10.27	& 10.5 \\
	& zFLoRA	& 23.03	& 35.75	& 61.3	& 116.16	& 248.93 & 7.6	& 7.62	& 7.69	& 7.78	& 7.97 \\
    \end{tabular}
    \caption{Latency measurements (in ms) made using vLLM inference engine on an NVIDIA H100 80GB GPU.}
    \label{tab:vllm}
\end{table*}
\begin{figure*}[t]
    \centering    
        \includegraphics[width=\linewidth]{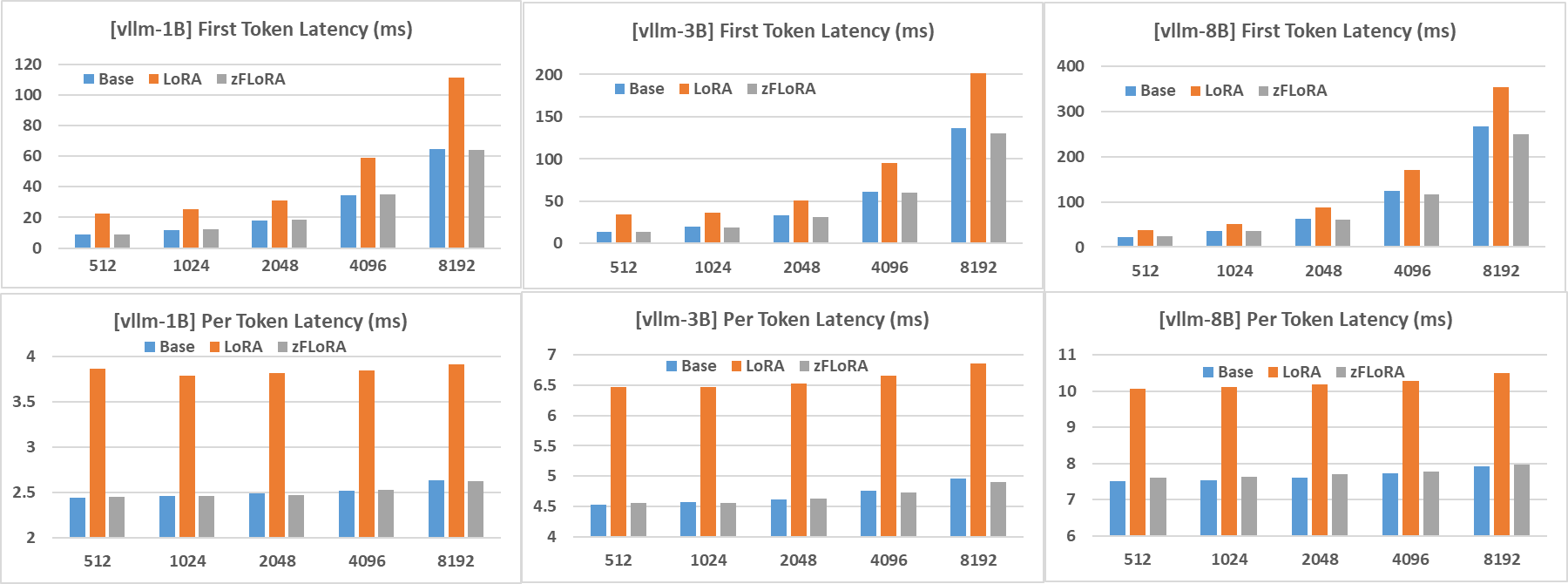}
    \vspace{-3mm}
    \caption{Inference latencies (first-token and per-token) in ms of base models (LLaMA3.x 1B, 3B and 8B) without and with adapters LoRA and zFLoRA for different input prompt lengths (512 to 2048) using vllm inference engine on NVIDIA H100 GPU at FP16 precision.}
    \label{fig:vllmlatms}
    \vspace{-5mm}
\end{figure*}
\section{vLLM inference latencies on H100 GPU (in ms)}
\label{app:vllm}
The detailed results of the latencies measured on an H100 GPU using vLLM inference engine in ms is given in Table~\ref{tab:vllm} and Fig.~\ref{fig:vllmlatms}.
The median and P99 ($99^{th}$ percentile) latencies have a similar trend and are not tabulated here.

\section{Detailed on-device latency measurements in ms}
\label{app:s25}
\begin{figure}[t]
    \centering    
        \includegraphics[width=\linewidth]{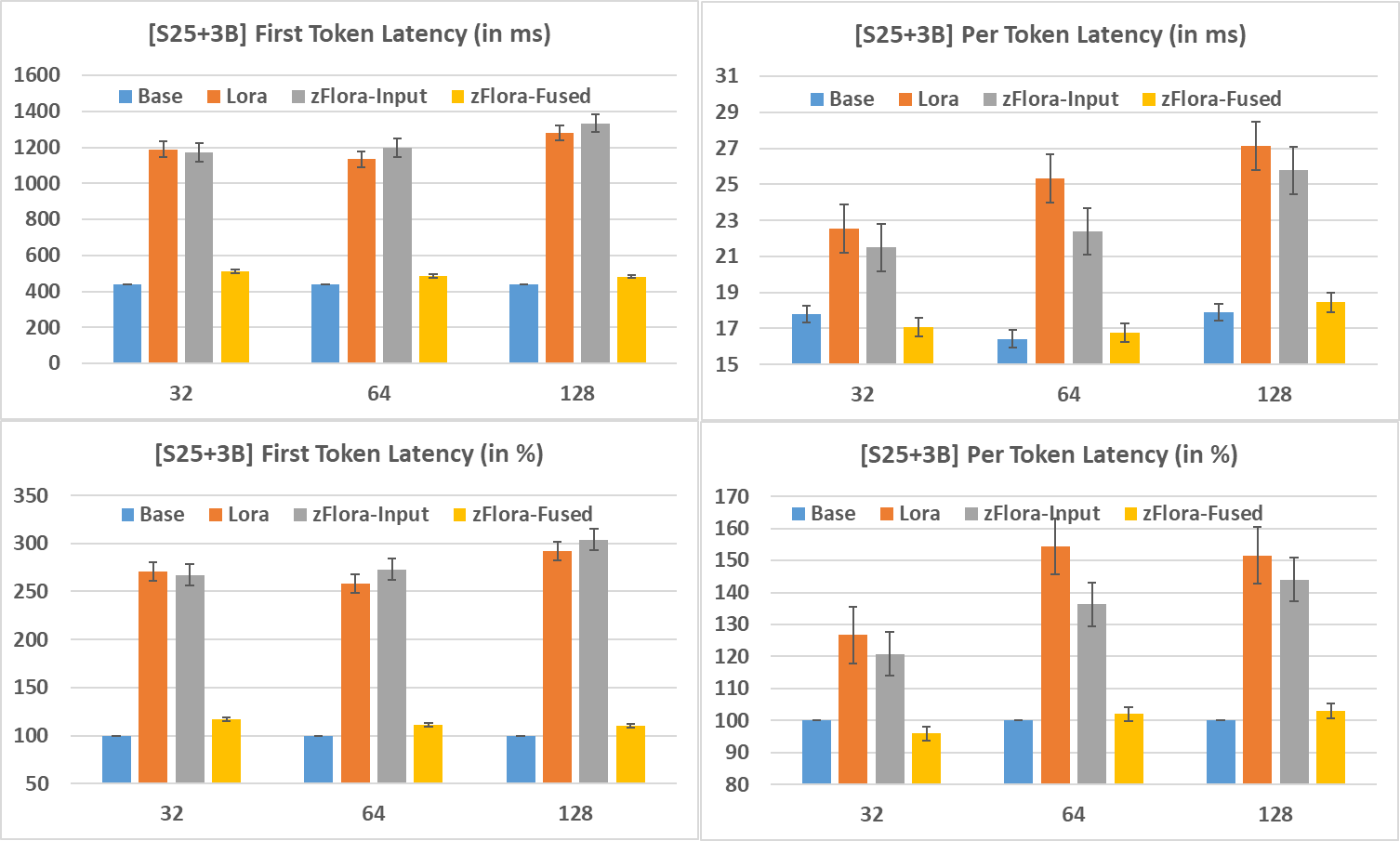}
    \vspace{-3mm}
    \caption{Inference latencies measured on Samsung Galaxy S25+ mobile handset for a 3B model.}
    \label{fig:s25-3b}
    \vspace{-5mm}
\end{figure}

The actual on-device latencies (in ms) measured on a Samsung Galaxy S25+ mobile handset with Qualcomm Snapdragon 8 Elite NPU chipset is given in Table~\ref{tab:s25} for different context lengths (with rank 32) and adapter ranks (with context length 1024).
For 3B model, latencies were measured only for varying ranks and corresponding plots are shown in Fig~\ref{fig:s25-3b}.

\begin{table*}[t]
    \centering
    {
    \begin{tabular}{c|rrr|rrr}
        \hline        
        & \multicolumn{3}{c|}{Prefill/First-token} & \multicolumn{3}{c}{Decode/Per-token} \\\hline
        \multicolumn{7}{c}{\bf 1B model} \\\hline
        Context	& 512	& 1024	& 2048 & 512	& 1024	& 2048 \\\hline
        Base	& 65.5	& 163.4	& 772.2 & 17.7	& 16.4	& 17.9 \\
        Lora	& 218.2	& 517.7	& 1582.4 & 22.5	& 25.3	& 27.1 \\
        zFlora-I	& 251.2	& 547.7	& 1565.5 & 21.4	& 22.3	& 25.7 \\
        zFlora-F	& 72.1	& 176.7	& 656.1 & 17.0	& 16.7	& 18.4 \\\hline
        
        Rank    & 32	& 64	& 128 & 32	& 64	& 128 \\\hline
        Base	& 163.45	& 163.45	& 163.45 & 16.42	& 16.42	& 16.42\\
        Lora	& 517.79	& 537.37	& 554.17 & 25.34	& 30.14	& 34.95 \\
        zFlora-I	& 547.75	& 594.43	& 640.64 & 22.38	& 28.19	& 30.12\\
        zFlora-F	& 176.7	& 185.7	& 184.02 & 16.75	& 18.93	& 18.39\\\hline
        \multicolumn{7}{c}{\bf 3B model} \\\hline
        & \multicolumn{3}{c}{Prefill/First-token} & \multicolumn{3}{c}{Decode/Per-token} \\\cline{2-7}
        Rank    & 32	& 64	& 128 & 32	& 64	& 128 \\\hline
        Base	& 438.5	& 438.5	& 438.5 & 17.7	& 16.4	& 17.9\\
        Lora	& 1188.7	& 1133.9	& 1280.1 & 22.5	& 25.3	& 27.1\\
        zFlora-I	& 1172.5	& 1197.6	& 1333.3 & 21.4	& 22.3	& 25.7\\
        zFlora-F	& 512.8	& 486.9	& 482.2 & 17.0	& 16.7	& 18.4
    \end{tabular}
    }
    \caption{S25+ on-device latencies (in ms) for a 1B/3B model for different context length and adapter ranks at W4A16 precision. zFLoRA-I and zFLoRA-F refer to zFLoRA-Input (input to graph) and zFLoRA-Fused (fused to the base model weights).}
    \label{tab:s25}
\end{table*}

\section{Performance of LoRA and zFLoRA for different ranks}
\label{app:perfvsrank}
Detailed performance of the LLaMA 1B-Inst and 3B-Inst models with LoRA and zFLoRA adapters for varying ranks is shown in Tables ~\ref{tab:perfvsrank1B} and ~\ref{tab:perfvsrank3B}.
Experiments for all 3 category of tasks were carried out for zFLoRA for both 1B and 3B model size.
Some math reasoning and summary-dialogue experiments were left out for the LoRA-3B combination, and may be conducted only if required.
The best LR obtained by coarse-and-fine LR sweeping for rank 32 was used for all other ranks.

\section{Performance of different fused-adapter variants}
\label{app:ffaffba}
The performance of FFA and FFBA adapters as compared to LoRA and zFLoRA adapters is given in Tables ~\ref{tab:ffaffba1B} and ~\ref{tab:ffaffba3B}.
As hypothesized earlier, it can be seen that the performance of FFA is inferior to other adapters which utilize LRA.
The FFBA (QG-Add) is a variant of the FFBA where forward adapters are attached only to query and gate projections, with the matching backward projections attached to MHA-output and FFN-down projection layers.
This eliminates the need for multiple merge operations on key, value and up projection layers.
It can be seen that FFBA (QG-Add) performs much better than FFA and closer to zFLoRA.
The FP32 latencies measured on an H100 GPU (averaged over 200 cnndm test utterances) show that FFA and FFBA adapters indeed reduce the latency overhead compared to LoRA but the additional merge or add operations introduce significant overheads as compared to zFLoRA.
zFLoRA (minimal) denotes the variant proposed in this paper as shown in Fig.~\ref{fig:zflora}, which uses minimal forward and backward adapter blocks.
zFLoRA (uniform) denotes another variant of zFLoRA that can also provide zero to negligible latencies, with both a forward and backward adapter attached to each layer in the transformer layer.
This leads to a uniform hidden dimension of $d+r$ throughout all layers of the model with an initial expansion and a final merging.
However, this increase in dimension leads to modifying the RoPE embeddings which is detrimental to the information learned by the pretrained LLM.
This leads to the poor convergence or performance of this zFLoRA (uniform) as can be seen the figure.
The modified architecture of zFLoRA (uniform) may need a few steps of uptraining (or continual pretraining) in order to address this issue, but is not investigated in this paper.

\section{Ablation experiment to reduce the adapter blocks}
\label{sec:ablation}
In the previous sections, the ablation experiments focused on studying the effect of rank size and the importance of forward and backward adapter blocks.
In both the cases, adapter blocks were attached to both the MHA and FFN blocks.
In this section, we study the possibility of reducing the overall adapter footprint by attaching the adapter blocks only to the MHA block.
In the case of zFLoRA, the backward adapters attached to the QKV layers as well as the forward adapter attached to the FFN down-projection layer are retained.
The experimental results are shown in Table~\ref{tab:onlymha}.
It can be seen that performance of both LoRA and zFLoRA degrade when adapters are attached only to the MHA block as compared to attaching them to both MHA and FFN blocks.
The degradation is less in the case of commonsense reasoning tasks which predict a single token.
However, in the case of math reasoning the degradation appears to be a bit more severe owing to the longer reasoning required.
zFLoRA appears to recover some lost performance as your increase the parameter count by increasing the adapter rank, a bit more gracefully compared to LoRA.
One possible reason for this behavior could be the cross-layer or across-the-block flow of information between the forward and backward adapters.
Nevertheless, when it comes to reducing the overall adapter footprint it may be better to attach adapters to both MHA and FFN blocks and reduce the rank as against attaching adapters only to the MHA block.
The other ablations of using the adapters only with the FFN blocks or with only a few selected transformer layers (top, bottom, mid, interleaved) can also be investigated, but not presented in this paper.

\begin{table*}
    \centering
    {
    \begin{tabular}{c|c|c|c|c}
        \hline        
        \multirow{2}{*}{\bf 1B-Inst} & & & \multicolumn{2}{c}{Common Sense Reasoning (acc)}  \\\cline{4-5}
        & Rank & \#Param & arcc~|~arce~|~boolq~|~hella~|~obqa~|~piqa~|~siqa~|~wino & Avg \\\hline
        Base & 0 & 1B & 51.00 | 73.00 | 64.00 | 44.00 | 74.50 | 72.50 | 50.00 | 45.00 & 59.25 \\\hline
        FFT	& 0	& 0 & 64.50 | 78.70 | 84.10 | 76.30 | 87.20 | 77.80 | 72.40 | 69.60 & 76.32 \\\hline
        & 4 & 2.8M & 61.80 | 77.10 | 76.50 | 73.10 | 80.40 | 75.10 | 72.00 | 65.60 & 72.70 \\
        & 8 &	5.6M &	62.00 | 78.20 | 81.70 | 76.30 | 86.20 | 78.80 | 71.80 | 69.90 &	75.61 \\
        LoRA & 16 &	11.2M &	64.50 | 80.00 | 82.50 | 75.90 | 85.40 | 77.40 | 73.10 | 69.70 &	76.06 \\
        (LR 5e-4) & 32 & 22.5M & 63.90 | 78.60 | 82.30 | 76.00 | 86.40 | 77.50 | 75.50 | 69.10 & 76.16 \\
        & 64 & 45M & 61.70 | 76.00 | 83.90 | 75.50 | 84.40 | 77.30 | 72.60 | 70.80 & 75.27 \\\hline
        & 4 &	1.9M &	64.00 | 76.70 | 78.90 | 76.20 | 82.00 | 74.30 | 72.40 | 68.40 &	74.11 \\
        & 8 & 3.8M &	62.20 | 77.50 | 78.60 | 75.10 | 85.00 | 77.00 | 71.80 | 68.90 &	74.51 \\
        zFLoRA & 16 & 7.6M & 62.10 | 77.60 | 81.80 | 76.10 | 85.00 | 77.10 | 72.40 | 68.30 & 75.05 \\
        (LR 2e-4) & 32 & 15.2M & 62.80 | 78.40 | 82.60 | 76.90 | 87.40 | 77.30 | 73.10 | 70.10 & 76.07 \\
        & 64 & 30.4M & 62.60 | 77.60 | 80.40 | 76.70 | 86.40 | 78.10 | 74.20 | 70.30 & 75.78 \\\hline       

        \multirow{2}{*}{\bf 1B-Inst} & & & \multicolumn{2}{c}{Math Reasoning (acc)} \\\cline{4-5}
        & Rank & \#Param & addsub~|~aqua~|~multi~|~gsm8k~|~singeq~|~svamp & Avg  \\\hline
        Base & 0 & 1B & 68.10 | 22.83 | 62.17 | 45.49 | 80.91 | 53.20 &	55.45 \\\hline
        FFT	& 0	& 0 & 85.32 | 22.83 | 96.17 | 48.52 | 90.94 | 66.70 & 68.41 \\\hline
        & 4 & 2.8M & 68.10 | 25.59 | 82.67 | 43.37 | 79.72 | 60.70 &  60.02 \\
        & 8 & 5.6M &	80.51 | 20.08 | 88.67 | 46.40 | 88.58 | 65.60 &	64.97 \\
        LoRA & 16 &	11.2M &	77.47 | 22.05 | 84.33 | 44.58 | 86.02 | 64.20 &	63.1 \\
        (LR 1e-4) & 32 & 22.5M & 82.78 | 28.35 | 92.67 | 48.14 | 87.99 | 67.00 & 67.82 \\
        & 64 & 45M & 75.19 | 24.41 | 86.67 | 45.19 | 82.09 | 59.70 & 62.2	\\\hline
        & 4 &	1.9M &	79.75 | 27.95 | 86.50 | 43.82 | 86.22 | 62.50 &	64.45 \\
        & 8 & 3.8M &	78.23 | 22.83 | 81.33 | 41.70 | 86.42 | 66.30 &	62.8 \\
        zFLoRA & 16 & 7.6M & 80.51 | 24.41 | 87.83 | 43.29 | 87.01 | 65.70 & 64.79 \\
        (LR 5e-4) & 32 & 15.2M & 87.85 | 24.80 | 96.00 | 43.37 | 91.93 | 59.40 & 67.22 \\
        & 64 & 30.4M & 89.62 | 23.62 | 95.83 | 39.80 | 91.14 | 61.50 & 66.91 \\\hline 

        \multirow{2}{*}{\bf 1B-Inst} & & & \multicolumn{2}{c}{Summary-Dialogue (RLsum)} \\\cline{4-5}
        & Rank & \#Param & cnndm~|~dd~|~woz~|~xsum & Avg \\\hline
        Base & 0 & 1B & 25.28 | 13.03 | 13.81 | 19.49 &	17.90 \\\hline
        FFT	& 0	& 0 & 28.37 | 16.58 | 30.45 | 32.67 & 27.01 \\\hline
        & 4 & 2.8M & 26.45 | 17.50 | 30.24 | 29.06 & 25.81 \\
        & 8 &	5.6M &	26.65 | 18.00 | 30.09 | 29.68 & 26.10 \\
        LoRA & 16 &	11.2M &	25.95 | 17.00 | 28.39 | 28.40 &	24.93 \\
        (LR 3e-4) & 32 & 22.5M & 26.76 | 20.12 | 31.34 | 32.23 &	27.61 \\
        & 64 & 45M & 27.24 | 17.67 | 29.95 | 31.75 &	26.65 \\\hline
        & 4 &	1.9M & 27.11 | 16.18 | 29.81 | 29.46 &	25.64 \\
        & 8 & 3.8M &	27.32 | 16.31 | 30.41 | 28.94 &	25.74 \\
        zFLoRA & 16 & 7.6M & 26.81 | 18.23 | 30.71 | 28.89 &	26.16 \\
        (LR 2e-4) & 32 & 15.2M & 27.25 | 18.31 | 31.82 | 30.98 &	27.09 \\
        & 64 & 30.4M & 27.37 | 19.73 | 32.54 | 31.32 & 27.74 \\\hline
    \end{tabular}
    }
    \caption{Performance of LLaMA 1B-Inst model with LoRA and zFLoRA adapters for varying ranks.}
    \label{tab:perfvsrank1B}
\end{table*}

\begin{table*}
    \centering
    {
    \begin{tabular}{c|c|c|c|c}
        \hline        
        \multirow{2}{*}{\bf 3B-Inst} & & & \multicolumn{2}{c}{Common Sense Reasoning (acc)}  \\\cline{4-5}
        & Rank & \#Param & arcc~|~arce~|~boolq~|~hella~|~obqa~|~piqa~|~siqa~|~wino & Avg \\\hline
        Base & 0 & 3B & 79.00 | 83.00 | 83.00 | 68.00 | 83.00 | 72.50 | 68.50 | 54.00 & 73.87 \\\hline
        FFT	& 0 & 0 & 79.00 | 86.40 | 89.30 | 85.40 | 93.20 | 84.70 | 80.40 | 83.20 & 85.2 \\\hline
        & r=4 & 6.1M & 77.00 | 87.30 | 88.00 | 84.10 | 91.80 | 84.70 | 81.60 | 82.90 & 84.67 \\
	& r=8 &	12.2M & 77.80 | 86.80 | 89.80 | 84.80 | 92.00 | 85.30 | 80.60 | 82.40 & 84.93 \\
	LoRA & r=16 & 24.3M & 77.10 | 86.60 | 90.00 | 86.00 | 93.20 | 85.40 | 80.10 | 83.70 & 85.26 \\
	(LR 5e-4)&r=32 & 48.6M & 77.60 | 86.00 | 89.20 | 84.90 | 93.00 | 85.40 | 80.80 | 84.50 & 85.17 \\
	&r=64 &	97.2M & 76.90 | 86.30 | 89.70 | 86.00 | 93.80 | 85.70 | 80.20 | 84.30 & 85.36 \\
	&r=128 & 194.4M & 78.10 | 87.10 | 88.70 | 86.30 | 92.00 | 84.70 | 80.90 | 84.50 & 85.28 \\\hline
        & r=4 & 3.6M & 77.00 | 86.70 | 87.10 | 83.70 | 90.40 | 82.30 | 79.50 | 79.90 & 83.32 \\
        &	r=8 & 7.2M & 77.60 | 85.90 | 87.80 | 84.40 | 90.60 | 83.00 | 79.50 | 82.30 & 83.88 \\
	zFLoRA & r=16 & 14.4M & 76.40 | 86.40 | 88.10 | 85.20 | 92.40 | 83.30 | 79.80 | 82.80 & 84.3 \\
	(LR 1e-4) & r=32 & 29M & 78.20 | 88.20 | 88.10 | 86.10 | 94.00 | 82.70 | 80.70 | 83.60 & 85.2 \\
	& r=64 & 59M & 76.90 | 87.90 | 89.40 | 84.40 | 92.80 | 85.30 | 79.90 | 84.50 & 85.13 \\
	& r=128 & 117M & 75.80 | 85.70 | 89.90 | 87.80 | 92.80 | 83.40 | 79.10 | 83.00 & 84.68 \\\hline

    \multirow{2}{*}{\bf 3B-Inst} & & & \multicolumn{2}{c}{Math Reasoning (acc)}  \\\cline{4-5}
        & Rank & \#Param & addsub~|~aqua~|~multi~|~gsm8k~|~singeq~|~svamp & Avg \\\hline
        Base & 0 & 3B & 91.14 | 24.80 | 93.17 | 76.88 | 93.90 | 87.60 & 77.91 \\\hline
        FFT	& 0 & 0 & 89.62 | 28.74 | 99.00 | 71.87 | 93.70 | 82.00 & 77.48 \\\hline
        & r=4 & 6.1M & -  & - \\
	& r=8 &	12.2M & - & - \\
	LoRA & r=16 & 24.3M & - & - \\
	(LR 3e-4) & r=32 & 48.6M & 93.16 | 27.17 | 96.67 | 67.10 | 95.87 | 82.50 & 77.07 \\
	&r=64 &	97.2M & - & - \\
	&r=128 & 194.4M & - & - \\\hline
        & r=4 & 3.6M & 91.14 | 29.53 | 98.17 | 67.78 | 94.69 | 77.40 & 76.45 \\
        &	r=8 & 7.2M & 88.86 | 25.98 | 97.00 | 68.39 | 92.13 | 80.00 & 75.39 \\
	zFLoRA & r=16 & 14.4M & 90.13 | 33.86 | 97.67 | 67.55 | 95.08 | 72.50 & 76.13 \\
	(LR 3e-4) & r=32 & 29M & 90.38 | 29.53 | 97.17 | 70.74 | 93.70 | 81.90 & 77.23 \\
	& r=64 & 59M & 89.62 | 26.38 | 95.67 | 70.89 | 95.28 | 81.50 & 76.55 \\
	& r=128 & 117M & 93.16 | 24.02 | 97.00 | 67.63 | 95.08 | 80.70 & 76.26 \\\hline

    \multirow{2}{*}{\bf 3B-Inst} & & & \multicolumn{2}{c}{Summary-Dialogue (RLsum)}  \\\cline{4-5}
        & Rank & \#Param & cnndm~|~dd~|~woz~|~xsum & Avg \\\hline
        Base & 0 & 3B & 91.14 | 24.80 | 93.17 | 76.88 | 93.90 | 87.60 & 77.91 \\\hline
        FFT	& 0 & 0 & 89.62 | 28.74 | 99.00 | 71.87 | 93.70 | 82.00 & 77.48 \\\hline
        & r=4 & 6.1M & - & - \\
	& r=8 &	12.2M & - & - \\
	LoRA & r=16 & 24.3M & - & - \\
	(LR 3e-5)&r=32 & 48.6M & 28.92 | 18.37 | 31.15 | 36.45 & 28.72 \\
	&r=64 &	97.2M & - & -\\
	&r=128 & 194.4M & - & \\\hline
        & r=4 & 3.6M & 28.13 | 16.81 | 28.78 | 32.21 & 26.48 \\
        &	r=8 & 7.2M & 27.41 | 17.19 | 31.97 | 33.26 & 27.45 \\
	zFLoRA & r=16 & 14.4M & 27.61 | 19.25 | 31.47 | 34.63 & 28.24 \\
	(LR 5e-5) & r=32 & 29M & 28.83 | 19.44 | 30.76 | 36.18 & 28.80  \\
	& r=64 & 59M & 27.38 | 19.20 | 31.76 | 36.38 & 28.68  \\
	& r=128 & 117M & 27.66 | 19.85 | 31.35 | 35.39 & 28.56 \\\hline
    \end{tabular}
    }
    \caption{Performance of LLaMA 3B-Inst model with LoRA and zFLoRA adapters for varying ranks.}
    \label{tab:perfvsrank3B}
\end{table*}

\begin{table*}
    \centering
    {
    \begin{tabular}{c|c|c|c|c|c}
    \hline  
    \multicolumn{6}{c}{\bf LLaMA 1B-Inst} \\\hline
    &  \multicolumn{5}{c}{Common Sense Reasoning (acc)} \\\cline{2-6}
    Adapter & \multicolumn{4}{c|}{arcc~|~arce~|~boolq~|~hella~|~obqa~|~piqa~|~siqa~|~wino} & Avg \\\hline
    Base & \multicolumn{4}{c|}{51.00 | 73.00 | 64.00 | 44.00 | 74.50 | 72.50 | 50.00 | 45.00} & 59.25  \\
    FFT & \multicolumn{4}{c|}{64.50 | 78.70 | 84.10 | 76.30 | 87.20 | 77.80 | 72.40 | 69.60} & 76.32  \\
    Lora & \multicolumn{4}{c|}{63.90 | 78.60 | 82.30 | 76.00 | 86.40 | 77.50 | 75.50 | 69.10} & 76.16 \\\hline
    FFA & \multicolumn{4}{c|}{52.50 | 71.00 | 81.50 | 69.50 | 85.00 | 69.50 | 69.50 | 69.50} & 71.00 \\
    FFBA (QG-Add) & \multicolumn{4}{c|}{62.10 | 76.00 | 79.90 | 73.40 | 84.60 | 77.70 | 71.70 | 68.90} & 74.28 \\
    zFLoRA (uniform) & \multicolumn{4}{c|}{(Poor performance due to RoPE modification)} & -  \\
    zFLoRA (minimal) & \multicolumn{4}{c|}{62.80 | 78.40 | 82.60 | 76.90 | 87.40 | 77.30 | 73.10 | 70.10} & 76.07 \\\hline\hline

    & \multicolumn{5}{c}{Math Reasoning (acc)} \\\cline{2-6}
    Adapter & \multicolumn{4}{c|}{addsub~|~aqua~|~multi~|~gsm8k~|~singeq~|~svamp} & Avg \\\hline
    Base & \multicolumn{4}{c|}{68.10 | 22.83 | 62.17 | 45.49 | 80.91 | 53.20} & 55.45  \\
    FFT & \multicolumn{4}{c|}{85.32 | 22.83 | 96.17 | 48.52 | 90.94 | 66.70} & 68.41  \\
    Lora & \multicolumn{4}{c|}{82.78 | 28.35 | 92.67 | 48.14 | 87.99 | 67.00}	& 67.82 \\\hline
    FFA & \multicolumn{4}{c|}{81.77 | 20.08 | 85.17 | 36.24 | 84.84 | 58.60} & 61.11  \\
    FFBA (QG-Add) & \multicolumn{4}{c|}{84.30 | 23.62 | 93.83 | 45.87 | 89.76 | 65.40} & 67.13 \\
    zFLoRA (uniform) & \multicolumn{4}{c|}{01.01 | 00.00 | 04.17 | 02.65 | 01.38 | 04.50} & 2.28 \\
    zFLoRA (minimal) & \multicolumn{4}{c|}{87.85 | 24.80 | 96.00 | 43.37 | 91.93 | 59.40} & 67.22 \\\hline\hline

    & & \multicolumn{2}{c|}{Latency} & \multicolumn{2}{c}{Summary-Dialogue (RLsum)} \\\cline{3-6}
    Adapter & Params & TTFT & TPOT & cnndm~|~dd~|~woz~|~xsum & Avg \\\hline
    Base & 1B & 11.9 & 6.6 & 25.28 | 13.03 | 13.81 | 19.49 & 17.9 \\
    FFT	& -	& - &	- &	28.37 | 16.58 | 30.45 | 32.67 & 27.01 \\
    Lora & 22.5M & 15.5 &	8.9 &	26.76 | 20.12 | 31.34 | 32.23 & 27.61 \\\hline
    FFA & 21M & 15.1 & 7.9 & 25.05 | 14.93 | 24.53 | 24.38 & 22.22 \\
    FFBA (QG-Add) & 21M & 14.7 & 8.2 & 26.24 | 19.67 | 29.65 | 29.38 & 26.23 \\
    zFLoRA (uniform) & 22.5M & 14 & 6.7 & 15.15 | 09.70 | 22.25 | 14.25 & 15.33 \\
    zFLoRA (minimal) & 15.2M & 13.2 & 6.5 & 27.25 | 18.31 | 31.82 | 30.98 & 27.09 \\\hline
    \end{tabular}
    }
    \caption{Performance of LLaMA 1B-Inst model for different fused adapter variants.}
    \label{tab:ffaffba1B}
\end{table*}

\begin{table*}
    \centering
    {
    \begin{tabular}{c|c|c|c|c|c}
    \hline  
    \multicolumn{6}{c}{\bf LLaMA 3B-Inst} \\\hline
    &  \multicolumn{5}{c}{Common Sense Reasoning (acc)} \\\cline{2-6}
    Adapter & \multicolumn{4}{c|}{arcc~|~arce~|~boolq~|~hella~|~obqa~|~piqa~|~siqa~|~wino} & Avg \\\hline
    Base & \multicolumn{4}{c|}{79.00 | 83.00 | 83.00 | 68.00 | 83.00 | 72.50 | 68.50 | 54.00} & 73.87  \\
    FFT & \multicolumn{4}{c|}{79.00 | 86.40 | 89.30 | 85.40 | 93.20 | 84.70 | 80.40 | 83.20} & 85.2  \\
    Lora & \multicolumn{4}{c|}{77.60 | 86.00 | 89.20 | 84.90 | 93.00 | 85.40 | 80.80 | 84.50} & 85.17 \\\hline
    FFA & \multicolumn{4}{c|}{76.00 | 84.50 | 85.00 | 78.00 | 88.50 | 76.00 | 78.50 | 77.50} & 80.5 \\
    FFBA (QG-Add) & \multicolumn{4}{c|}{77.60 | 86.60 | 88.00 | 85.40 | 92.20 | 83.70 | 78.70 | 83.10} & 84.41 \\
    zFLoRA (uniform) & \multicolumn{4}{c|}{(Poor performance due to RoPE modification)} & - \\
    zFLoRA (minimal) & \multicolumn{4}{c|}{78.20 | 88.20 | 88.10 | 86.10 | 94.00 | 82.70 | 80.70 | 83.60} & 85.2 \\\hline\hline

    & \multicolumn{5}{c}{Math Reasoning (acc)} \\\cline{2-6}
    Adapter & \multicolumn{4}{c|}{addsub~|~aqua~|~multi~|~gsm8k~|~singeq~|~svamp} & Avg \\\hline
    Base & \multicolumn{4}{c|}{91.14 | 24.80 | 93.17 | 76.88 | 93.90 | 87.60} & 77.91 \\
    FFT & \multicolumn{4}{c|}{89.62 | 28.74 | 99.00 | 71.87 | 93.70 | 82.00} & 77.48 \\
    Lora & \multicolumn{4}{c|}{93.16 | 27.17 | 96.67 | 67.10 | 95.87 | 82.50} & 77.07 \\\hline
    FFA & \multicolumn{4}{c|}{87.59 | 21.26 | 96.00 | 66.87 | 92.13 | 80.30} & 74.02 \\
    FFBA (QG-Add) & \multicolumn{4}{c|}{90.13 | 33.86 | 97.33 | 69.45 | 94.88 | 80.00} & 77.6 \\
    zFLoRA (uniform) & \multicolumn{4}{c|}{(Poor performance due to RoPE modification)} & - \\
    zFLoRA (minimal) & \multicolumn{4}{c|}{90.38 | 29.53 | 97.17 | 70.74 | 93.70 | 81.90} & 77.23 \\\hline\hline

    & & \multicolumn{2}{c|}{Latency} & \multicolumn{2}{c}{Summary-Dialogue (RLsum)} \\\cline{3-6}
    Adapter & Params & TTFT & TPOT & cnndm~|~dd~|~woz~|~xsum & Avg \\\hline
    Base & 3B & 25.5 & 11.7 & 25.10 | 14.45 | 16.68 | 20.54 & 19.19 \\
    FFT	& - & - & - & 29.23 | 25.85 | 29.66 | 37.63 & 30.59 \\
    Lora & 48.6M & 31.9 & 15.2 & 28.92 | 18.37 | 31.15 | 36.45 & 28.72 \\\hline
    FFA & 55M & 30.6 & 13.2 & 26.04 | 18.45 | 28.67 | 31.85 & 26.25 \\
    FFBA (QG-Add) & 55M & 30.5 & 13.5 & 28.71 | 20.39 | 30.87 | 35.72 & 28.92 \\
    zFLoRA (uniform) & 55M & 30.9 & 11.6 & 13.69 | 04.54 | 19.00 | 15.03 & 13.06 \\
    zFLoRA (minimal) & 29.3M & 28 & 10.9 & 28.83 | 19.44 | 30.76 | 36.18 & 28.8 \\\hline
    \end{tabular}
    }
    \caption{Performance of LLaMA 3B-Inst model for different fused adapter variants.}
    \label{tab:ffaffba3B}
\end{table*}

\begin{table*}
    \centering
    {
    \begin{tabular}{c|c|c|c|c}
        \hline        
        \multirow{2}{*}{\bf 1B-Inst} & & & \multicolumn{2}{c}{Common Sense Reasoning (acc)}  \\\cline{4-5}
        & Rank & \#Param & arcc~|~arce~|~boolq~|~hella~|~obqa~|~piqa~|~siqa~|~wino & Avg \\\hline
        Base & 0 & 1B & 51.00 | 73.00 | 64.00 | 44.00 | 74.50 | 72.50 | 50.00 | 45.00 & 59.25 \\\hline
        FFT	& 0	& 0 & 64.50 | 78.70 | 84.10 | 76.30 | 87.20 | 77.80 | 72.40 | 69.60 & 76.32 \\\hline
        LoRA-MHA & 4 & 0.8M &  58.60 | 74.80 | 74.80 | 69.70 | 77.00 | 71.80 | 68.20 | 60.30 & 69.40 \\
        (LR 5e-4) & 32 & 6.8M & 61.90 | 76.90 | 81.80 | 74.60 | 86.20 | 74.00 | 71.90 | 69.10 & 74.55 \\
        & 64 & 13.6M & 62.10 | 75.40 | 81.60 | 75.00 | 86.00 | 76.50 | 71.30 | 69.90 & 74.72 \\\hline
        zFLoRA-MHA & 4 & 0.7M &	59.20 | 75.00 | 77.30 | 71.70 | 80.20 | 74.60 | 69.20 | 62.20 & 71.17 \\
        (LR 2e-4) & 32 & 5.7M & 58.50 | 76.50 | 76.40 | 71.40 | 80.80 | 75.00 | 70.40 | 62.60 & 71.45 \\
        & 64 & 11.5M & 62.50 | 75.40 | 81.00 | 75.10 | 85.40 | 76.90 | 72.50 | 68.70 & 74.68 \\\hline\hline
        \multirow{2}{*}{\bf 1B-Inst} & & & \multicolumn{2}{c}{Math Reasoning (acc)} \\\cline{4-5}
        & Rank & \#Param & addsub~|~aqua~|~multi~|~gsm8k~|~singeq~|~svamp & Avg  \\\hline
        Base & 0 & 1B & 68.10 | 22.83 | 62.17 | 45.49 | 80.91 | 53.20 &	55.45 \\\hline
        FFT	& 0	& 0 & 85.32 | 22.83 | 96.17 | 48.52 | 90.94 | 66.70 & 68.41 \\\hline
        LoRA-MHA & 4 & 0.8M & 67.85 | 25.20 | 69.50 | 41.70 | 76.77 | 57.70 & 56.45 \\
        (LR 1e-4) & 32 & 6.8M & 65.82 | 22.44 | 75.00 | 43.06 | 75.98 | 55.70 & 56.33 \\
        & 64 & 13.6M & 58.73 | 24.02 | 79.83 | 42.15 | 74.41 | 53.30 & 55.40 \\\hline
        zFLoRA-MHA & 4 & 0.7M &	63.04 | 23.23 | 79.17 | 42.46 | 72.24 | 56.30 & 56.07 \\
        (LR 5e-4) & 32 & 5.7M & 69.11 | 23.23 | 81.00 | 41.70 | 78.15 | 63.50 & 59.44 \\
        & 64 & 11.5M & 85.57 | 27.17 | 94.17 | 44.66 | 88.78 | 67.60 & 67.99 \\\hline 
    \end{tabular}
    }
    \caption{Performance of LLaMA 1B-Inst model when adapters are attached only to the MHA block.}
    \label{tab:onlymha}
\end{table*}

\end{document}